\documentclass{article}
\PassOptionsToPackage{numbers, compress}{natbib}
\usepackage[preprint]{neurips22}

\usepackage[ruled]{algorithm2e}
\usepackage{bbm}
\usepackage{amsmath}
\usepackage{amsfonts}
\usepackage{multirow}
\usepackage{array}
\usepackage[pdftex]{graphicx}

\usepackage[utf8]{inputenc} 
\usepackage[T1]{fontenc}    
\usepackage{hyperref}       
\usepackage{url}            
\usepackage{booktabs}       
\usepackage{amsfonts}       
\usepackage{nicefrac}       
\usepackage{microtype}      
\usepackage{xcolor}         
\usepackage{ulem}           

\title{Multivariate Prediction Intervals for Random Forests}

\author{%
    Brendan Folie \\
    Citrine Informatics \\
    \texttt{bfolie@citrine.io} \\
    \And
    Maxwell Hutchinson \\
    Kiva Microfunds\footnotemark \\
    \texttt{maxh@kiva.org} \\
    }

\begin{document}

\maketitle

\footnotetext{Maxwell Hutchinson also performed the work while at Citrine Informatics.}

\begin{abstract}
Accurate uncertainty estimates can significantly improve the performance of iterative design of experiments, as in Sequential and Reinforcement learning.
For many such problems in engineering and the physical sciences, the design task depends on multiple correlated model outputs as objectives and/or constraints.
To better solve these problems, we propose a recalibrated bootstrap method to generate multivariate prediction intervals for bagged models and show that it is well-calibrated. We apply the recalibrated bootstrap to a simulated sequential learning problem with multiple objectives and show that it leads to a marked decrease in the number of iterations required to find a satisfactory candidate. This indicates that the recalibrated bootstrap could be a valuable tool for practitioners using machine learning to optimize systems with multiple competing targets.
\end{abstract}

\section{Introduction}

One drawback of many regression algorithms is that they do not naturally admit an uncertainty estimate. It is often important to know not just what a model predicts but how confident it is in that prediction. For situations in which a machine learning model is making a single decision, especially one of great import to humans, such as who gets a mortgage or who is granted parole \citep{Kuchibhotla:21}, an overly-confident model can be disastrous. Accurate uncertainty estimates are also crucial when applying machine learning to the physical sciences. For example, a practitioner may be using a model to identify a material that achieves outstanding performance in some application \citep{Meredig:18}. Many models are unable to predict values substantially outside the range of the training data, so a direct prediction of such performance is unlikely. But with a well-calibrated uncertainty estimate we may be able to identify candidates that have a reasonable probability of achieving high performance. For this reason uncertainty estimates are an integral part of sequential learning \citep{Ling:17}, an iterative method that seeks to identify configurations that meet some ambitious targets. Sequential learning has been used to identify novel organic LED materials \citep{Abroshan:21,Antono:20}, battery materials \citep{Verduzco:21,Dave:20} and charging protocols \citep{Attia:20}, colloidal nanoparticles \citep{Fong:21}, perovskite photovoltaics \citep{Liu:22} and many more.

Extending this idea to multiple outputs, a model should be able to produce multivariate prediction intervals that identify the correlations between outputs. This is relevant to sequential learning and reinforcement learning, which often seek to optimize properties that are in tension with each other. For example, a scientist may seek to create a new thermoelectric material that efficiently converts waste heat into electricity. This requires high electrical conductivity and low thermal conductivity, but increasing one tends to increase the other. Or a rover may seek to plan a trajectory through space that maximizes the information it collects over the trajectory while minimizing the length of the trajectory \citep{Wang:18}. In these cases it is not enough to know that a proposed candidate has a 20\% chance of satisfying each objective individually. Depending on the correlation this candidate could be highly promising or utterly useless.

Here we propose a \emph{multi-output recalibrated bootstrap} method to generate prediction intervals for bagged models by rescaling the bootstrap standard deviation based on the out-of-bag errors. This method is fast, easy to implement, and works with small numbers of bags and training rows. We study the accuracy of the recalibrated bootstrap on a variety of synthetic and real-world test problems, showing that it is well-calibrated in both the single- and multi-output cases. Finally we test the recalibrated bootstrap on a multi-output sequential learning scenario, in which the algorithm attempts to identify a candidate that will simultaneously satisfy several competing objectives. We find a marked decrease in the number of iterations required, which has significant implications for real-world sequential learning.

\section{Related Work}

Several early and influential studies of uncertainty in random forests focused on generating a confidence interval for the predicted value. \citet{Wager:14} studied two generalizations of the jackknife, the infinitesimal jackknife and the jackknife after bootstrap, and introduced a bias-correction term that decreases the number of bags needed for the calculation to converge. Around the same time \citet{Mentch:16} proposed a modification to the sub-sampling procedure that allows for the generation of confidence intervals. These procedures estimate model variability: by how much is the prediction likely to vary if the model were retrained on a new set of training data drawn from the underlying distribution? \textbf{This is not the same as a prediction interval}, which is an attempt to bound the true value. To drive this distinction home, Figure \ref{fig:jackknife_vs_error} in Appendix \ref{apx:jackknife_underestimates_error} plots the jackknife standard deviation and model error for two one-dimensional test functions and shows that the jackknife can greatly underestimate the true prediction error. Despite this mismatch many works have used the jackknife methods of \citet{Wager:14} as if they were a prediction interval \citep{Roman:21,Wahab:20,Ruesch:20,Carrella:21,Ling:17,Lepioufle:21}, sometimes leading to sub-optimal results. For example, when using the infinitesimal jackknife to estimate uncertainty as part of a sequential learning study, \citet{Rohr:20} take note of the ``...general overconfidence of [jackknife-based] methods.''

So how does one generate a prediction interval? Quantile Regression Forests (QRF) \citep{Meinshausen:06} were an early attempt. A separate model is trained for each desired quantile. QRFs are highly versatile but require large amounts of data to train and can be noisy \citep{Zhang:20}. The last few years have seen renewed interest in the subject. \citet{Lei:18} developed a general ``split conformal'' method for producing regression prediction intervals. Conformal methods are flexible in that they make no assumptions about the model. They generate prediction intervals by examining the residuals on new test points. The split conformal method reserves some training data for this purpose, hence it is data-inefficient. \citet{Dewolf:21} conduct a review of methods to generate prediction intervals on regression problems and find that different methods are better at different problems, but all methods can achieve high accuracy if a conformal method is used to recalibrate the prediction intervals.

\citet{Zhang:20} specialized the conformal approach to random forest and made it more efficient by considering the out-of-bag (OOB) residuals. The OOB predictions are those made by base learners that were not exposed to a given training point, and hence they provide a reasonable proxy for how the model will perform on new data points. \citet{Zhang:20} prove that their OOB prediction interval has an asymptotically correct coverage rate given certain assumptions.
But one drawback of this approach is that the interval is not conditioned on the input, meaning it does not generalize well when the noise is heteroskedastic or the training and test data are not drawn from the same underlying distribution. The lack of conditioning is particularly problematic when designing an experiment (e.g. sequential learning), where the prediction interval should be smaller near previously sampled regions of the domain and larger in unexplored ones.

A conditional prediction interval is proposed by \citet{Lu:21}, who weight the OOB residuals based on how similar each training point is to the test point. Similarity is determined by the tree structure -- a test point is similar to training points with which it shares the leaf node. In this way, the prediction interval is sensitive to the local OOB error.
Another conditional prediction interval is suggested by \citet{Palmer:21}, who estimate the local uncertainty by calculating the standard deviation over the base learner predictions.
This value is shifted and rescaled by constant factors that are estimated based on the residuals computed by cross-validation.

As far as we can tell, the question of multivariate prediction intervals for bagged learners is entirely unexplored in the literature.

\section{Recalibrated Bootstrap Prediction Intervals}

Consider $N$ training examples $Z_i = (\vec{x}_i, \vec{y}_i)$, where $\vec{x}$ is an input vector and $\vec{y}$ is an output vector. The inputs can be of any type but the outputs are real-valued. Assume that the training data are drawn according to some ground-truth function $f(\vec{x})$ with sampling noise parameterized by a distribution $\epsilon(\vec{x})$. That is, $\vec{y}_i = f(\vec{x}) + \epsilon(\vec{x})$.

We draw $B$ bootstrap samples of the training data, each of which is generated by sampling $N$ times with replacement. The resulting sample is used to train a base learner that makes predictions $\vec{t}_b(\vec{x})$. The model prediction is the mean of the base learners: $\hat{\vec{\theta}}(\vec{x}) = \frac{1}{B} \sum_{b=1}^B \vec{t}_b(\vec{x})$. We use a carat to denote a quantity that is estimated over the bootstrap samples.

The recalibrated bootstrap technique proposed here generates prediction intervals that are given by a normal distribution centered on $\hat{\vec{\theta}}(\vec{x})$. The only thing that needs to be determined therefore is the covariance matrix of the normal distribution. We take this to be the covariance matrix of the bootstrap predictions, but we rescale each entry using recalibration factors that are determined using the OOB predictions. We use the notation $(-i)$ to determine a quantity computed over the OOB trees for training point $i$.

We first consider a single response $y$ and show how to generate a standard deviation that corresponds to a normal prediction interval. First we must choose a confidence level parameter $p \in (0, 1)$ and calculate the equivalent number of standard deviations of a normal distribution: $\eta(p) = \Phi^{-1}(\frac{1 + p}{2})$ where $\Phi$ is the CDF of a unit normal distribution. Given $p$, Algorithm \ref{alg:recalibrated_bootstrap} describes how to calculate the recalibration factor $\alpha$ and apply it to new predictions.

\begin{algorithm}\label{alg:recalibrated_bootstrap}
 \caption{Recalibrated Bootstrap Prediction Interval in One Dimension}\label{alg:recalibrated_bootstrap}
\SetAlgoLined
\textbf{Stage} : Determine recalibration factor $\alpha_p$\;
\textbf{Given} : $N$ training points $(\vec{x}_i, y_i)$\;
\textbf{Given} : $B$ bootstrap samples and base learners $t_b(\vec{x})$\;
\textbf{Given} : A confidence level parameter, $p \in (0, 1)$\;
\For{all training points $\vec{x}_i$}{
    Calculate the out-of-bag mean, $\hat{\theta}_{(-i)}(\vec{x})$\;
    Calculate the out-of-bag standard deviation, $\hat{s}_{(-i)}(\vec{x})$\;
    Calculate the out-of-bag standard residual, $|\tilde{r}_{oob}| = \displaystyle\frac{|\hat{\theta}_{(-i)}(\vec{x}) - y_i|}{\hat{s}_{(-i)}}$\;
}
Let $\tilde{r}_p$ be the $p$-percentile value of the $|\tilde{r}_{oob}|$\;
Set $\alpha$ equal to $\tilde{r}_p / \eta(p)$, where $\eta(p)$ is the equivalent number of standard deviations of a normal distribution;

\textbf{Stage} : Estimate prediction interval for a new input\;
\textbf{Given} : new input $\vec{x}$\;
Calculate $\hat{s}(\vec{x})$, the standard deviation over the predictions $t(\vec{x})$\;
The prediction interval is defined by a normal distribution with mean $\hat{\theta}(\vec{x})$ and standard deviation $\alpha * \hat{s}(\vec{x})$
\end{algorithm}

Like \citet{Palmer:21}, this method computes factors to reclaibrate the bootstrap standard deviation. But there are two key differences. First, using the OOB residuals instead of cross-validation makes the method more data-efficient. Second, as we will show in the subsequent analysis, using the $p$-percentile instead of the maximum likelihood estimate (MLE) makes the method more robust to outliers as long as the chosen value of $p$ is not too extreme.

The generalization to a multivariate prediction interval is straight-forward. For each pair of outputs we calculate a recalibrated covariance using the individual recalibration factors. Let $\alpha_j$ and $\alpha_k$ be the recalibration factors for outputs $j$ and $k$. For a given output $j$ let $t_{bj}(\vec{x})$ be the prediction of tree $b$, $\hat{\theta}_j(\vec{x})$ be the mean prediction, and $\hat{\sigma_j}(\vec{x})$, written explicitly in Equation \ref{eqn:recalibrated_variance}, be the recalibrated standard deviation that forms the univariate prediction interval.

\begin{equation}\label{eqn:recalibrated_variance}
\hat{\sigma}_j(\vec{x}) = \alpha_j \sqrt{\frac{1}{B-1} \sum_{b=1}^B (t_{bj}(\vec{x}) - \hat{\theta_j}(\vec{x}))^2}
\end{equation}

The covariance estimate is then given by Equation \ref{eqn:recalibrated_covariance}.

\begin{equation}\label{eqn:recalibrated_covariance}
\begin{aligned}
\hat{\sigma}^2_{jk} &= \frac{1}{B - 1} \sum_{b=1}^B \Big( \alpha_j (t_{bj}(\vec{x}) - \hat{\theta}_j(\vec{x})) \Big) \Big( \alpha_k (t_{bk}(\vec{x}) - \hat{\theta}_k(\vec{x})) \Big) \\
& = \frac{\displaystyle\sum_b \Big(t_{bj}(\vec{x}) - \hat{\theta_j}(\vec{x})\Big) \Big(t_{bk}(\vec{x}) - \hat{\theta_k}(\vec{x})\Big)}{\sqrt{\Big(\displaystyle\sum_b (t_{bj}(\vec{x}) - \hat{\theta_j}(\vec{x}))^2 \Big) \Big(\sum_b (t_{bk}(\vec{x}) - \hat{\theta_k}(\vec{x}))^2\Big)}} \hat{\sigma}_j(\vec{x}) \hat{\sigma}_k(\vec{x}) \\
& \equiv \hat{\rho}_{jk}(\vec{x}) \hat{\sigma}_{j}(\vec{x}) \hat{\sigma}_k(\vec{x})
\end{aligned}
\end{equation}

The bootstrap correlation coefficient $\hat{\rho}_{jk}(\vec{x})$ is shown to be the ordinary Pearson correlation coefficient calculated over the tree-wise predictions. If the resulting covariance matrix is $\hat{\Sigma}(\vec{x})$ then the prediction interval region is defined by a normal distribution $\mathcal{N}(\hat{\vec{\theta}}(\vec{x}), \hat{\Sigma}(\vec{x}))$. We refer to this distribution as the \emph{prediction distribution}.

This is a pleasing result -- by forcing the prediction distribution to be normal and by using the tree-wise standard deviation as a proxy for uncertainty, we have arrived at a multivariate distribution that is simple to generate and to evaluate, making it ideal for multi-objective sequential learning. Other methods to compute a prediction interval, such as \citet{Zhang:20} or \citet{Lu:21}, are more flexible but do not have a clear multivariate generalization, and any such generalization would likely require the amount of data to grow exponentially with the number of outputs.

The efficacy of this procedure will be established in subsequent section by showing that the prediction distribution is well calibrated and that it leads to more efficient sequential learning. But it is also worth briefly considering the theoretical justification for Algorithm \ref{alg:recalibrated_bootstrap}, which presupposes that the standard OOB residuals are drawn from the same distribution as the test set residuals and also that the bootstrap variance is proportional to the squared residual. We show in Appendices \ref{apx:residual_equivalence} and \ref{apx:recalibrated_bootstrap_captures_error} that these are reasonable propositions, given that the training and test sets are drawn independently from identical distributions. In real-world data sets, including those used in this work, the training and test sets are often drawn from non-identical distributions. But as we will see, the recalibrated bootstrap still performs well.

The choice of a single $p$ assumes that the distribution of standard residuals is normally distributed.
We investigate that normality numerically, by training a large number of random forest models on different training data draws and examining the distribution of standard OOB residuals. Results are shown in Figure \ref{fig:recalibration}a for the Friedman-Grosse function (see Appendix \ref{apx:Friedman_Grosse}). Considering both the PDF and the CDF we see that the values are evenly distributed and close to normal, but have slightly fatter tails. The inability to capture these tails is one drawback of restricting the prediction interval to be a normal distribution. The results on other test problems are similar (see Appendix \ref{apx:recalibration_factors}), though real-world data sets tend to produce a more skewed distribution.

If the distribution of standardized residuals is normal, then any $p$ can be used to generate intervals with any confidence level via post-hoc rescaling of the interval by $\eta(p)/\eta(p')$.
However, certain choices of $p$ may result in more reliable intervals, generally.
In Figure \ref{fig:recalibration}b we plot the recalibration factor vs. $p$. Ideally it would be uniform, but we see that it increases sharply around $p = 0.9$. This is consistent with the distribution having fat tails -- a larger recalibration factor is needed to capture the larger residuals. We also consider the recalibration factor that maximizes the total log likelihood of the OOB residuals, similar to what is proposed in \citet{Palmer:21}, and plot it as a dashed line. Maximizing the log likelihood proposes a larger recalibration factor because it is sensitive to the penalty imposed by underestimating the large residuals.

For simplicity and because we do not want to be overly skewed by outliers, we set $p = 0.683$ for the remainder of this manuscript. The impact of the value of $p$ on sequential learning performance is a potential topic for further study.

\begin{figure}
    \centering
    \includegraphics[width=0.85\linewidth]{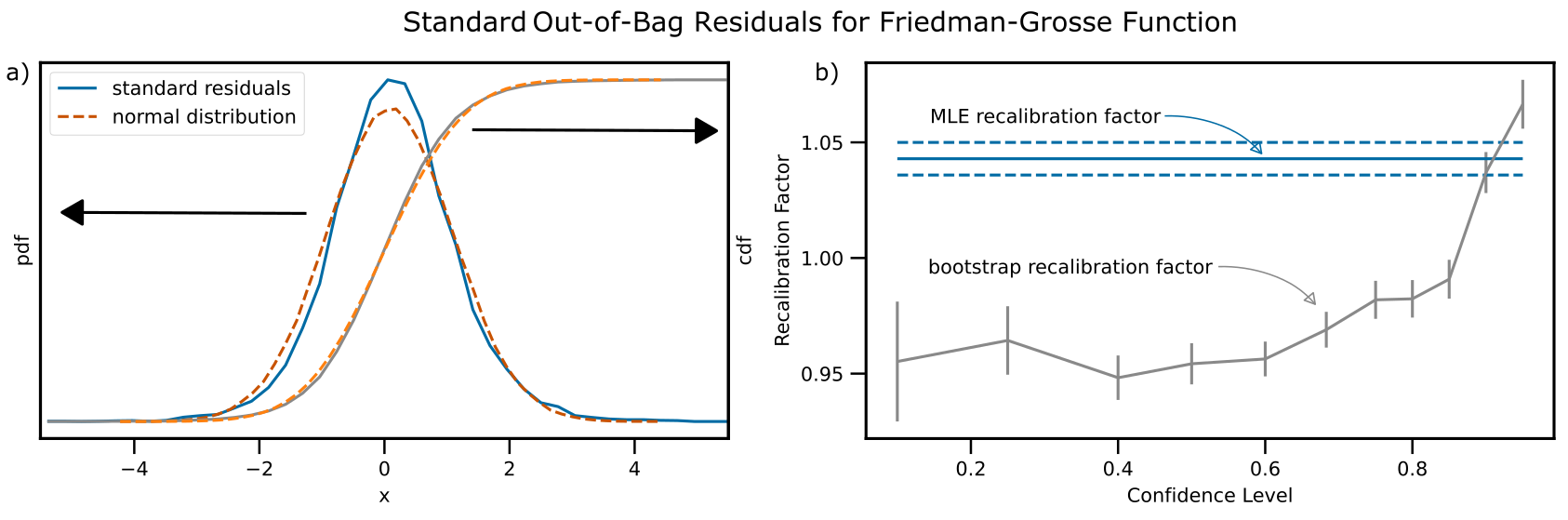}
    \caption{The standard out-of-bag (OOB) residuals approximate a normal distribution, hence the recalibrated bootstrap is expected to produce a well-calibrated prediction interval. This is seen by a) comparing the PDF and CDF to that of an ideal normal and b) checking the uniformity of the recalibration factor as a function of confidence level. We also calculate the recalibration factor generated using the MLE method and see that it is strongly influenced by outliers. Data were generated using the Friedman-Grosse function with a noise level of 2.0. There are 128 training points, 64 bags, and 100 trials.}
    \label{fig:recalibration}
\end{figure}

\section{Numerical Experiments: How Well Calibrated is the Proposed Prediction Interval?}

We perform numerical experiments to investigate the accuracy of the prediction intervals proposed by the recalibrated bootstrap method. We consider modifications of two commonly-used synthetic problems, Friedman-Silverman and Friedman-Grosse. We also consider two real-world data sets, ``thermoelectrics'' \citep{Gaultois:13} and ``mechanical properties.'' All of the data sets are dense, although the methods described here are amenable to sparse data as well. \citep{Borg:20} Details of the data sets are in Appendix \ref{apx:test_problems}.

For each test problem we generate training and test data sets, train a random forest model on the training data, apply the model to the test data, and evaluate the performance of the proposed prediction intervals by comparing them to the ``observed'' values. For all models we calculate the recalibration factor using $p = 0.683$, corresponding to $\eta \approx 1$.

All random forest models were trained with the package Lolo \citep{Hutchinson:16}, which is available under the Apache License 2.0. Unless otherwise stated, all forests contained 64 decision trees. All decision trees were grown to full depth and all inputs were considered at each split. The split was chosen to maximize the reduction in total variance summed over all outputs. All outputs were standardized before training to have mean 0 and variance 1.

All of the figures and numerical results reported in this work, along with the code and data used to generate those results, are available at \url{https://github.com/CitrineInformatics/multivariate-prediction-intervals}. All numerical experiments were run on a Mac Book Pro with a 2.3 GHz Quad-Core Intel Core i5 processor. Most experiments ran in 1 minute or less. The exception is simulated sequential learning, which took about 2 hours on the synthetic data and about 50 hours on the thermoelectrics data.

\subsection{Metrics}
We consider three metrics of prediction interval quality.

\textbf{Standard Error}: the mean of the absolute residual divided by the predicted uncertainty, as given in Equation \ref{eqn:standard_error} (the sum is over the test data). 
This value should be around 1.0, and it should be stable as the number of training data are varied. This metric is only used for univariate prediction intervals.

\begin{equation}\label{eqn:standard_error}
    \texttt{Standard Error} = \frac{1}{M} \sum_j \frac{|\hat{\theta}(\vec{x}_j) - y_j|}{\hat{\sigma}(\vec{x}_j)}
\end{equation}

\textbf{Standard Confidence}: how closely does the number of residuals within some magnitude match the number of residuals that are \textit{expected} to be within that magnitude? The magnitude of the residual is the Mahalanobis distance, $r_M = \sqrt{\vec{r}^T\hat{\Sigma}(\vec{x}_j)^{-1} \vec{r}}$, where $\vec{r} = \hat{\vec{\theta}}(\vec{x}_j) - \vec{y}_j$ is the residual. The squared Mahalanobis distance follows a $\chi^2$ distribution with $d$ degrees of freedom, where $d$ is the number of output dimensions. For a given coverage level $p_c \in (0, 1)$ we can therefore use the inverse CDF of $\chi^2_d$ to calculate the associated cutoff distance $\eta_c$, and count the number of observations for which $r_M < \eta_c$. In one dimension this reduces to Equation \ref{eqn:standrad_confidence}

\begin{equation}\label{eqn:standrad_confidence}
    \texttt{Standard Confidence} = \frac{1}{M} \sum_j \mathbbm{1} \Big[ \frac{|\hat{\theta}(\vec{x}_j) - y_j|}{\hat{\sigma}(\vec{x}_j)} \leq \Phi^{-1} \Big( \frac{1 + p_c}{2} \Big) \Big]
\end{equation}

If the standard confidence is greater than $p_c$ then the model is under-confident, and if it is less than $p_c$ then the model is over-confident. In this work we use ``standard confidence'' to refer to the case when $p_c \approx 0.683$.

\textbf{Median Negative Log Probability Density (NLPD)}: Let $p(\vec{y}; \hat{\vec{\theta}}(\vec{x}), \hat{\Sigma}(\vec{x}))$ be the probability density function of the prediction distribution at point $\vec{x}$. The NLPD for test point $(\vec{x}_j, \vec{y}_j)$. is given by Equation \ref{eqn:nlpd}. Lower values are better. NLPD penalizes both over- and under-confident prediction intervals. Because NLPD is prone to outliers (especially for the jackknife method), here we consider the median value over all test points.

\begin{equation}\label{eqn:nlpd}
    \texttt{NLPD}(\vec{x}_j) = -\ln(p(\vec{y}_j; \hat{\vec{\theta}}(\vec{x}_j), \hat{\Sigma}(\vec{x}_j)))
\end{equation}

\subsection{Univariate Calibration}

\begin{figure}[ht]
    \centering
    \includegraphics[width=0.75\linewidth]{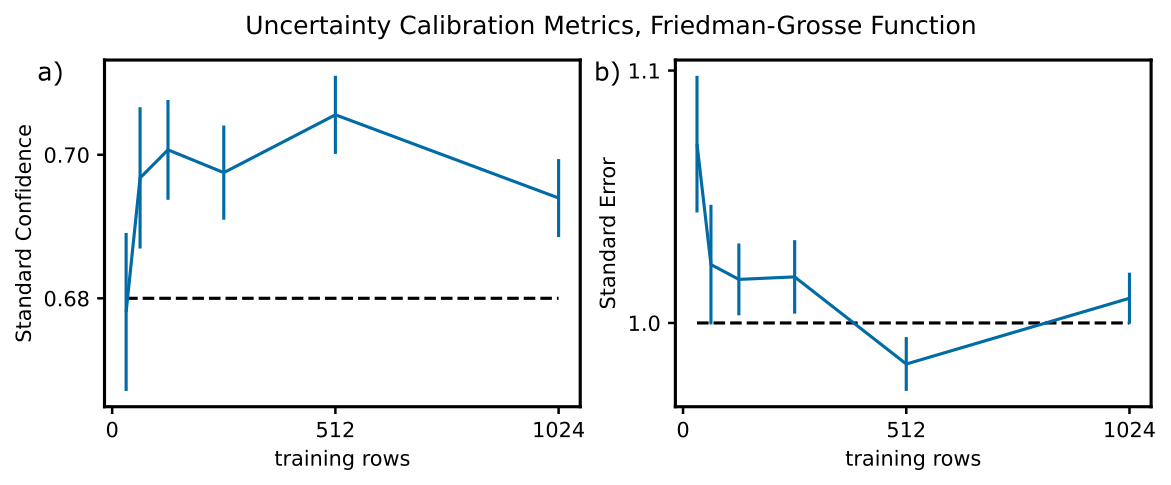}
    \caption{The recalibrated bootstrap produces a well-calibrated univariate prediction interval on synthetic data, as seen by examining a) standard confidence and b) standard error. Data are generated using the Friedman-Grosse function with normally distributed noise of magnitude 2.0. There are 128 test points and each random forest has 64 bags. 64 trials were run, each involving a different set of training and test data. Error bars show one standard error.}
    \label{fig:uncertainty_metrics}
\end{figure}

We show that the recalibrated bootstrap is well-calibrated by calculating the standard confidence and standard error for data generated using the Friedman-Grosse function. As shown in Figure \ref{fig:uncertainty_metrics} the standard confidence is close to 0.68 and the standard error is close to 1.0, as expected. In Figure \ref{fig:univariate_metrics_supplementary} we consider several other test problems and find that the standard confidence is consistently close to 0.68 but the standard error can be larger than 1 for real-world data. These two observations indicate that there are a few outliers for which the prediction interval is highly over-confident.

An in-depth comparison of univariate prediction intervals would consider several other techniques, such as those proposed in \citet{Palmer:21}, \citet{Zhang:20}, and \citet{Lu:21}. But this is not an in-depth comparison, nor do we assert that the recalibrated bootstrap is always the superior approach in one dimension. We merely show that it consistently produces well-calibrated univariate prediction intervals. In the subsequent sections we show that it also produces well-calibrated multivariate prediction intervals and that this quality results in significantly more efficient sequential learning. Using different techniques to produce the univariate prediction interval and studying the impact on SL would be a valuable topic for further study.

We also performed several sanity checks, some of which are included in the appendices. We consider the prediction interval quality when the test and training data are drawn from different distributions, finding that the recalibrated bootstrap becomes somewhat over-confident but is reasonably robust (Appendix \ref{apx:imbalanced_data}). We consider the impact of the number of bags, and find it is largely irrelevant (Figure \ref{fig:metrics_vary_bags}). And we consider the behavior of the recalibrated bootstrap in the high-noise limit, finding that it correctly identifies the noise as the primary source of uncertainty (Figure \ref{fig:metrics_vary_noise}).

\subsection{Multivariate Calibration}

To investigate the multivariate case we calculate $\hat{\sigma}_j(\vec{x})$ using the recalibrated bootstrap method and consider four different methods of calculating $\hat{\rho}_{jk}(\vec{x})$.

1. \textbf{Trivial}: $\hat{\rho}_{jk} = 0$, the probability of satisfying each objective is considered independent.

2. \textbf{Training Data}: Calculate the Pearson correlation coefficient over the training data and and use that value for all predictions.

3. \textbf{Jackknife}: Calculate the jackknife variance, $\text{V}_J$, and the jackknife covariance, $\text{Cov}_J$, and set $\hat{\rho}_{jk} = \text{Cov}_J[j, k] / \sqrt{\text{V}_J[j] * \text{V}_J[k]}$. See Appendix \ref{apx:jackknife} for more details.

4. \textbf{Bootstrap}: Calculate the Perason correlation coefficient over the tree-wise predictions as in Equation \ref{eqn:recalibrated_covariance}. This is the approach we are proposing here.

It might seem strange to consider the jackknife method, since it produces a quantity that is known to be more confident than a prediction interval. But it is possible that this bias will cancel out between the variance and covariance terms, leaving us with a decent estimate of correlation.

\begin{figure}[ht]
    \centering
    \includegraphics[width=0.85\linewidth]{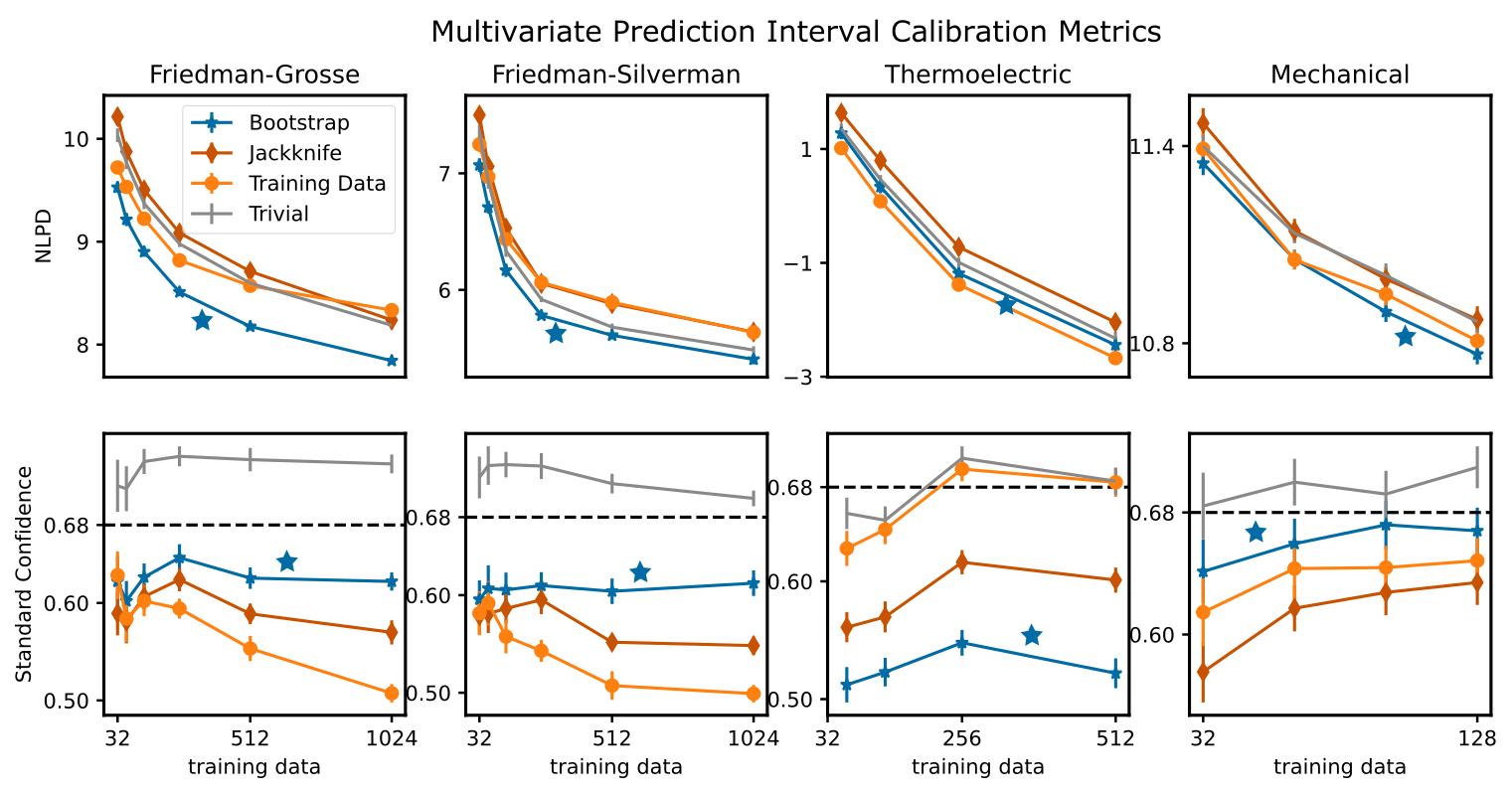}
    \caption{The recalibrated bootstrap produces the best overall multivariate prediction intervals on a suite of test problems, as judged by a low NLPD (top row) and a standard confidence that is close to 0.68 (bottom row). The four test problems are Friedman-Grosse with noise 2.0 (128 test points, 16 trials), Friedman-Silverman with noise 2.0 (128 test points, 16 trials), thermoelectrics (64 test points, 32 trials), and mechanical properties (48 test points, 32 trials). See Appendix \ref{apx:test_problems} for details. In all instances the random forest models have 64 bags.}
    \label{fig:covariance_metrics}
\end{figure}

Figure \ref{fig:covariance_metrics} shows the NLPD and Standard Confidence values for four test problems using each approach to calculating $\rho_{jk}$. A star icon is placed next to the curve for the bootstrap method, to draw attention to it. We see that the bootstrap method generally has the best NLPD and a standard confidence that is good but slightly under-confident. The trivial method is slightly over-confident by a similar amount. The thermoelectrics data set is an exception, for which the bootstrap method is significantly under-confident. Yet as we will see in the next section, that does not prevent the recalibrated bootstrap from achieving superior performance on simulated sequential learning.

\section{Application to Multi-Objective Sequential Learning}

A lower NLPD value is all well and good, but what we really want to know is whether or not this prediction distribution can help us make better decisions. We consider sequential learning (SL), also known as active learning, as a test problem \citep{Ling:17}. A user attempts to find an input point $\vec{x}$ that will lead to output $\vec{y}$ that simultaneously satisfies their objectives. Given initial training data, we train a multi-output random forest model. The model makes predictions at all unknown points, and an acquisition function is used to evaluate the suitability of some test point. The highest-scoring test point is chosen for measurement. If it satisfies the objectives then we are done. If not then it is added to the training set and the cycle repeats. This is similar in spirit to Bayesian Optimization \citep{Shahriari:16}.

The choice of the acquisition function depends on the problem at hand.
Here we use the predicted probability of the candidate lying in the satisfactory region.
This probability is estimated by drawing 10,000 samples from the prediction distribution. In other situations it might be preferable to use other acquisition functions, such as the probability of being Pareto non-dominated \citep{delRosario:20}.

We simulate SL on two problems. First, we modify the Friedman-Grosse function to generate two outputs that have a non-trivial relationship to each other. The outputs are positively correlated in one region and negatively correlated in another region. See Appendix \ref{apx:Friedman_Grosse} for more details. The objectives are that each output should exceed 22, a value chosen to make the problem challenging. As seen in Figure \ref{fig:synthetic_data} of Appendix \ref{apx:Friedman_Grosse}, only two points satisfy both objectives.

Second, we consider the thermoelectrics data set and devise a problem for which there is one solution: ZT $>$ 1.25, Seebeck coefficient $>$ 175 $\mu \text{V}/\text{K}$, power factor $>$ $5 \times 10^{-3} \text{ Wm}/\text{K}^2$, and thermal conductivity $>$ 1.5 $\text{W}/\text{mK}$. The data are plotted in Figure \ref{fig:thermoelectrics_data} of Appendix \ref{apx:thermoelectrics}.

\begin{figure}[ht]
    \centering
    \includegraphics[width=0.75\linewidth]{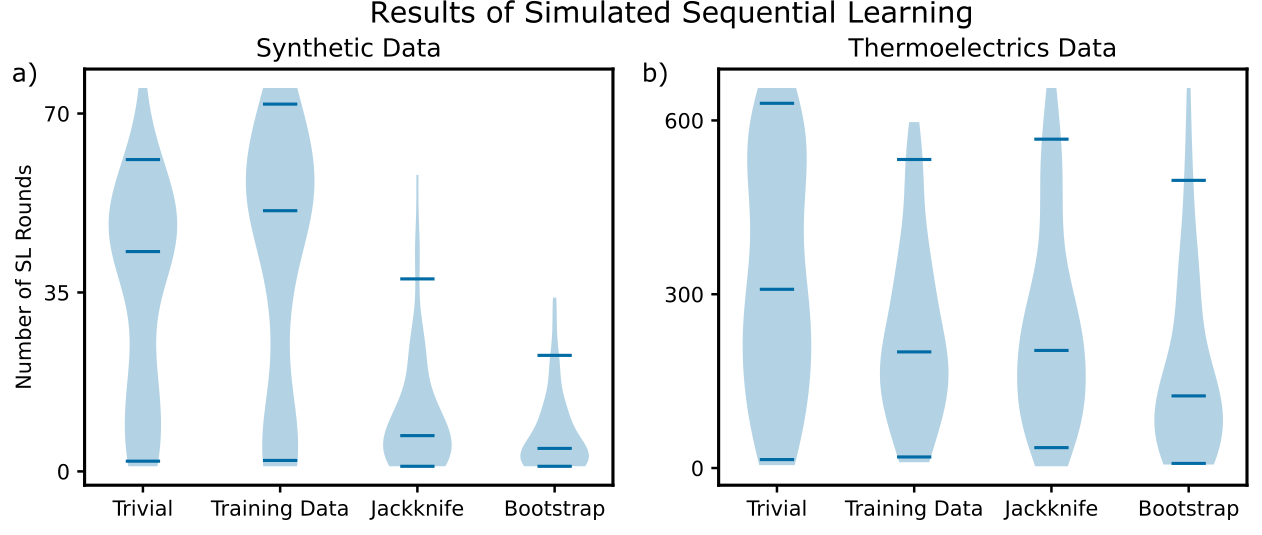}
    \caption{Using the recalibrated bootstrap decreases the number of iterations required to find a configuration that satisfies multiple competing objectives. It is better than the jackknife method and much better than the trivial method, which ignores correlations. The underlying data for the a) synthetic and b) thermoelectrics problem are plotted along with dashed lines to indicate the satisfactory regions. Violin plots in c) and d) show the distribution of results over 64 trials, along with lines to indicate the 5th percentile, median, and 95th percentile. The synthetic data problem starts with 16 training data and the thermoelectrics problem starts with 32 training data.}
    \label{fig:sequential_learning}
\end{figure}

\begin{table}
\caption{Distribution of rounds required to find a satisfactory candidate in simulated sequential learning. Results are calculated over 64 independent trials.}
\label{tbl:sequential_learning}
\centering
\begin{tabular}{c m{6em} c c c c} 
 \toprule
   & Method & mean & 5th percentile & median & 95th percentile \\
 \midrule
 \multirow{4}{7em}{Synthetic Data\\(2 objectives)} & Trivial & 36.0 $\pm$ 2.6 & 2 & 43 & 61 \\ 
 & Training Data & 41.8 $\pm$ 3.1 & 2 & 51 & 72 \\ 
 & Jackknife & 11.2 $\pm$ 1.5 & \textbf{1} & 7 & 37.7 \\
 & \textbf{Bootstrap} & \textbf{7.5 $\pm$ 1.0} & \textbf{1} & \textbf{4.5} & \textbf{22.7} \\
 & Random & 36 & 3 & 33 & 87\\
 \midrule
 \multirow{4}{7em}{Thermoelectrics Data\\(4 objectives)} & Trivial & 330 $\pm$ 25 & 15 & 309 & 630 \\
 & Training Data & 230 $\pm$ 19 & 19 & 200.5 & 532.5\\
 & Jackknife & 240 $\pm$ 21 & 35 & 203 & 568\\
 & \textbf{Bootstrap} & \textbf{170 $\pm$ 20} & \textbf{8} & \textbf{125} & \textbf{497} \\
 & Random & 328 & 33 & 328 & 623\\
 \bottomrule
\end{tabular}
\end{table}

Figure \ref{fig:sequential_learning} shows the results of 64 trials of simulated SL. Each trial involved a different set of initial training data points. The distribution of rounds to find a satisfactory candidate is shown as a violin plot. The 5th percentile, median, and 95th percentile values are indicated with horizontal lines. These values, along with the mean and standard error, are shown in Table \ref{tbl:sequential_learning}. Table \ref{tbl:sequential_learning} also includes the values that would be expected if test points were drawn randomly. The bootstrap consistently requires the fewest iterations to find a satisfactory candidate. Compared to the trivial method, the median trial requires 90\% fewer iterations for the synthetic problem and 60\% fewer iterations for the thermoelectrics problem. Perhaps more importantly, the trivial method suffers from a long tail of disastrous trials in which the performance is worse than that of random guessing. The recalibrated bootstrap largely avoids this pathology.

\section{Conclusions}

We propose a ``multi-output recalibrated bootstrap'' method to generate multivariate prediction intervals for random forest or any other bagged ensemble model. The covariance matrix of the bootstrap predictions defines a multivariate normal distribution, but the values are rescaled using the out-of-bag residuals. We show that this prediction interval is well-calibrated by testing it against several data sets.

Our focus is not, however, the prediction interval itself, but its applicability to multi-output sequential learning (SL).
We simulate SL on both synthetic and real data and show that using the recalibrated bootstrap significantly decreases the number of iterations required to find a high-performing candidate by between 2x and 8x compared to common alternatives.
Although this study is limited in that it only considers a few data sets, we feel that the dramatic improvement seen on real-world, noisy, imbalanced data (see the OOB residuals in Figure \ref{fig:recalibration_supplementary}) is promising enough to merit its use and further study.

We expect that the recalibrated bootstrap will be an invaluable tool for practitioners using machine learning to accelerate their technology development.
To date, SL has largely been used to identify novel materials for renewable energy applications, an endeavor with positive societal impact.
Developing the next-generation materials crucial to renewable energy generation and storage is an ideal problem: there are typically multiple ambitious goals that compete against each other, existing data sets are modest in size, and each experiment can be enormously expensive.
But SL is a value-agnostic algorithm for optimizing any complex system using machine learning, and it can just as easily be used to increase the lethality of explosives or the potency of a nicotine cartridge.
Our contribution increases the efficacy of SL, and hence its societal impact depends on the problems it is applied to.

\newpage

\bibliography{bibliography}

\begin{thebibliography}{37}
\providecommand{\natexlab}[1]{#1}
\providecommand{\url}[1]{\texttt{#1}}
\expandafter\ifx\csname urlstyle\endcsname\relax
  \providecommand{\doi}[1]{doi: #1}\else
  \providecommand{\doi}{doi: \begingroup \urlstyle{rm}\Url}\fi

\bibitem[Abroshan et~al.(2021)Abroshan, Chandrasekaran, Winget, An, Kwak,
  Brown, and Halls]{Abroshan:21}
Hadi Abroshan, Anand Chandrasekaran, Paul Winget, Yuling An, Shaun Kwak,
  Christopher Brown, and Mathew~D. Halls.
\newblock Accelerated design and optimization of novel oled materials via
  active learning.
\newblock \emph{SPIE Organic Photonics and Electronics}, 11808, 2021.

\bibitem[Antono et~al.(2020)Antono, Matsuzawa, Ling, Saal, Arai, Sasago, and
  Fujii]{Antono:20}
Erin Antono, Nobuyuki~N. Matsuzawa, Julia Ling, James~Edward Saal, Hideyuki
  Arai, Masaru Sasago, and Eiji Fujii.
\newblock Machine-learning guided quantum chemical and molecular dynamics
  calculations to design novel hole-conducting organic materials.
\newblock \emph{Journal of Physical Chemistry A}, 124\penalty0 (40):\penalty0
  8330--8340, 2020.

\bibitem[Attia et~al.(2020)Attia, Grover, Jin, Severson, Markov, Liao, Chen,
  Cheong, Perkins, Yang, Herring, Aykol, Harris, Braatz, Ermon, and
  Chueh]{Attia:20}
Peter~M. Attia, Aditya Grover, Norman Jin, Kristen~A. Severson, Todor~M.
  Markov, Yang-Hung Liao, Michael~H. Chen, Bryan Cheong, Nicholas Perkins,
  Zi~Yang, Patrick~K. Herring, Muratahan Aykol, Stephen~J. Harris, Richard~D.
  Braatz, Stefano Ermon, and William~C. Chueh.
\newblock Closed-loop optimization of fast-charging protocols for batteries
  with machine learning.
\newblock \emph{Nature}, 578:\penalty0 397--402, 2020.

\bibitem[Borg et~al.(2020)Borg, Frey, Moh, Pollock, Gorsse, Miracle, Senkov,
  Meredig, and Saal]{Borg:20}
Christopher K.~H. Borg, Carolina Frey, Jasper Moh, Tresa~M. Pollock, Stéphane
  Gorsse, Daniel~B. Miracle, Oleg~N. Senkov, Bryce Meredig, and James~E. Saal.
\newblock Expanded dataset of mechanical properties and observed phases of
  multi-principal element alloys.
\newblock \emph{Scientific Data}, 7, 2020.

\bibitem[Carrella(2021)]{Carrella:21}
Ernesto Carrella.
\newblock No free lunch when estimating simulation parameters.
\newblock \emph{The Journal of Artificial Societies and Social Simulations},
  24, 2021.

\bibitem[del Rosario et~al.(2020)del Rosario, Rupp, Kim, Antono, and
  Ling]{delRosario:20}
Zachary del Rosario, Matthias Rupp, Yoolhee Kim, Erin Antono, and Julia Ling.
\newblock Assessing the frontier: Active learning, model accuracy, and
  multi-objective candidate discovery and optimization.
\newblock \emph{The Journal of Chemical Physics}, 153, 2020.

\bibitem[Dewolf et~al.(2021)Dewolf, Baets, and Waegeman]{Dewolf:21}
Nicolas Dewolf, Bernard~De Baets, and Willem Waegeman.
\newblock Valid prediction intervals for regression problems.
\newblock \emph{ArXiv}, 2021.
\newblock \doi{arXiv:2017.00363v2}.

\bibitem[Efron(1982)]{Efron:82}
Bradley Efron.
\newblock The jackknife, the bootstrap, and other resampling plans.
\newblock \emph{The Society for Industrial and Applied Mathematics}, 1982.

\bibitem[Efron(2014)]{Efron:14}
Bradley Efron.
\newblock Estimation and accuracy after model selection.
\newblock \emph{Journal of the American Statistical Association}, 109, 2014.

\bibitem[Fong et~al.(2021)Fong, Pellouchoud, Davidson, Walroth, Church,
  Tcareva, Wu, Peterson, Meredig, and Tassone]{Fong:21}
Anthony~Y. Fong, Lenson Pellouchoud, Malcolm Davidson, Richard~C. Walroth,
  Carena Church, Ekaterina Tcareva, Liheng Wu, Kyle Peterson, Bryce Meredig,
  and Christopher~J. Tassone.
\newblock Utilization of machine learning to accelerate colloidal synthesis and
  discovery.
\newblock \emph{The Journal of Chemical Physics}, 154, 2021.

\bibitem[Friedman and Silverman(1989)]{Friedman:89}
J.~H. Friedman and B.~W. Silverman.
\newblock Flexible parsimonious smoothing and additive modeling.
\newblock \emph{Technometrics}, 3:\penalty0 3--39, 1989.

\bibitem[Friedman et~al.(1983)Friedman, Grosse, and Stuetzle]{Friedman:83}
J.~H. Friedman, E.~Grosse, and W.~Stuetzle.
\newblock Multidimensional additive spline approximation.
\newblock \emph{SIAM J. Sci. Statist. Comput.}, 4:\penalty0 291--301, 1983.

\bibitem[Gaultois et~al.(2013)Gaultois, Sparks, Borg, Seshadri, Bonificio, and
  Clarke]{Gaultois:13}
Michael~W. Gaultois, Taylor~D. Sparks, Christopher K.~H. Borg, Ram Seshadri,
  William~D. Bonificio, and David~R. Clarke.
\newblock Data-driven review of thermoelectric materials: Performance and
  resource considerations.
\newblock \emph{Chemistry of Materials}, 25\penalty0 (15):\penalty0 2911--2920,
  2013.

\bibitem[Ghosal(2021)]{Ghosal:21}
Indrayudh Ghosal.
\newblock \emph{Model Combinations and the Infinitesimal Jackknife: How to
  Refine Models With Boosting and Quantify Uncertainty}.
\newblock PhD thesis, Cornell University, 2021.

\bibitem[Hutchinson(2016)]{Hutchinson:16}
Maxwell Hutchinson.
\newblock Lolo.
\newblock \url{https://github.com/CitrineInformatics/lolo}, 2016.

\bibitem[Kandasamy et~al.(2020)Kandasamy, Burke, Paria, Póczos, Whitacre, and
  Viswanathan]{Dave:20}
Adarsh Dave Jared Mitchell~Kirthevasan Kandasamy, Han Wang~Sven Burke, Biswajit
  Paria, Barnabás Póczos, Jay Whitacre, and Venkatasubramanian Viswanathan.
\newblock Autonomous discovery of battery electrolytes with robotic
  experimentation and machine learning.
\newblock \emph{Cell Reports Physical Science}, 1, 2020.

\bibitem[Kuchibhotla and Berk(2021)]{Kuchibhotla:21}
Arun~K. Kuchibhotla and Richard~A. Berk.
\newblock Nested conformal prediction sets for classification with applications
  to probation data.
\newblock \emph{ArXiv}, 2021.
\newblock \doi{arXiv:2104.09358v1}.

\bibitem[Lei et~al.(2018)Lei, G'Sell, Rinaldo, Tibshirani, and
  Wasserman]{Lei:18}
Jing Lei, Max G'Sell, Alessandro Rinaldo, Ryan~J. Tibshirani, and Larry
  Wasserman.
\newblock Distribution-free predictive inference for regression.
\newblock \emph{Journal of the American Statistical Association}, 113:\penalty0
  1094--1111, 2018.

\bibitem[Lepioufle et~al.(2021)Lepioufle, Marsteen, and Johnsrud]{Lepioufle:21}
Jean-Marie Lepioufle, Leif Marsteen, and Mona Johnsrud.
\newblock Error prediction of air quality at monitoring stations using random
  forest in a total error framework.
\newblock \emph{Physical Sensors}, 21, 2021.

\bibitem[Ling et~al.(2017)Ling, Hutchinson, Antono, Paradiso, and
  Meredig]{Ling:17}
Julia Ling, Maxwell Hutchinson, Erin Antono, Sean Paradiso, and Bryce Meredig.
\newblock High-dimensional materials and process optimization using data-driven
  experimental design with well-calibrated uncertainty estimates.
\newblock \emph{Integrating Materials and Manufacturing Innovation},
  6:\penalty0 207--217, 2017.

\bibitem[Liu et~al.(2022)Liu, Rolston, Flick, Colburn, Ren, Dauskardt, and
  Buonassisi]{Liu:22}
Zhe Liu, Nicholas Rolston, Austin~C. Flick, Thomas~W. Colburn, Zekun Ren,
  Reinhold~H. Dauskardt, and Tonio Buonassisi.
\newblock Machine learning with knowledge constraints for process optimization
  of open-air perovskite solar cell manufacturing.
\newblock \emph{ArXiv}, 2022.
\newblock \doi{arXiv:2110.01387v4}.

\bibitem[Lu and Hardin(2021)]{Lu:21}
Benjamin Lu and Johanna Hardin.
\newblock A unified framework for random forest prediction error estimation.
\newblock \emph{Journal of Machine Learning Research}, 22:\penalty0 1--41,
  2021.

\bibitem[Meinshausen(2006)]{Meinshausen:06}
Nicolai Meinshausen.
\newblock Quantil regression forests.
\newblock \emph{Journal of Machine Learning Research}, 7:\penalty0 983--999,
  2006.

\bibitem[Mentch and Hooker(2016)]{Mentch:16}
Lucas Mentch and Giles Hooker.
\newblock Quantifying uncertainty in random forests via confidence intervals
  and hypothesis tests.
\newblock \emph{Journal of Machine Learning Research}, 17:\penalty0 1--41,
  2016.

\bibitem[Meredig et~al.(2018)Meredig, Antono, Church, Hutchinson, Ling,
  Paradiso, Blaiszik, Foster, Gibbons, Hattrick-Simpers, Mehta, and
  Ward]{Meredig:18}
Bryce Meredig, Erin Antono, Carena Church, Maxwell Hutchinson, Julia Ling, Sean
  Paradiso, Ben Blaiszik, Ian Foster, Brenna Gibbons, Jason Hattrick-Simpers,
  Apurva Mehta, and Logan Ward.
\newblock Can machine learning identify the next high-temperature
  superconductor? examining extrapolation performance for materials discovery.
\newblock \emph{Molecular Systems Design and Engineering}, 3:\penalty0 819 --
  825, 2018.

\bibitem[Palmer et~al.(2021)Palmer, Du, Politowicz, Emory, Yang, Gautam, Gupta,
  Li, Jacobs, and Morgan]{Palmer:21}
Glenn Palmer, Siqi Du, Alexander Politowicz, Joshua~Paul Emory, Xiyu Yang,
  Anupraas Gautam, Grishma Gupta, Zhelong Li, Ryan Jacobs, and Dane Morgan.
\newblock Calibrated bootstrap for uncertainty quantification in regression
  models.
\newblock \emph{ArXiv}, 2021.
\newblock \doi{arXiv:2105.13303v1}.

\bibitem[Rohr et~al.(2020)Rohr, Stein, Guevarra, Wang, Haber, Aykol, Suram, and
  Gregoire]{Rohr:20}
Brian Rohr, Helge~S. Stein, Dan Guevarra, Yu~Wang, Joel~A. Haber, Muratahan
  Aykol, Santosh~K. Suram, and John~M. Gregoire.
\newblock Benchmarking the acceleration of materials discovery by sequential
  learning.
\newblock \emph{Chemical Science}, 11:\penalty0 2696--2706, 2020.

\bibitem[Roman et~al.(2021)Roman, Saxena, Robu, Pecht, and Flynn]{Roman:21}
Darius Roman, Saurabh Saxena, Valentin Robu, Michael Pecht, and David Flynn.
\newblock Machine learning pipeline for battery state-of-health estimation.
\newblock \emph{Nature Machine Intelligence}, 3:\penalty0 447--456, 2021.

\bibitem[Ruesch et~al.(2020)Ruesch, Yang, Schmitt, Acharya, Smith, and
  Kainerstorfer]{Ruesch:20}
Alexander Ruesch, Jason Yang, Samantha Schmitt, Deepshikha Acharya, Matthew~A.
  Smith, and Jana~M. Kainerstorfer.
\newblock Estimating intracranial pressure using pulsatile cerebral blood flow
  measured with diffuse correlation spectroscopy.
\newblock \emph{Biomedical Optics Express}, 11\penalty0 (3), 2020.

\bibitem[Shahriari et~al.(2016)Shahriari, Swersky, Wang, Adams, and
  de~Freitas]{Shahriari:16}
Bobak Shahriari, Kevin Swersky, Ziyu Wang, Ryan~P. Adams, and Nando de~Freitas.
\newblock Taking the human out of the loop: A review of bayesian optimization.
\newblock \emph{Proceedings of the IEEE}, 104\penalty0 (1), 2016.

\bibitem[Verduzco et~al.(2021)Verduzco, Marinero, and Strachan]{Verduzco:21}
Juan~C. Verduzco, Ernesto~E. Marinero, and Alejandro Strachan.
\newblock An active learning approach for the design of doped llzo ceramic
  garnets for battery applications.
\newblock \emph{Integrating Materials and Manufacturing Innovation},
  10:\penalty0 299--310, 2021.

\bibitem[Wager et~al.(2014)Wager, Hastie, and Efron]{Wager:14}
Stefan Wager, Trevor Hastie, and Bradley Efron.
\newblock Confidence intervals for random forests: The jackknife and the
  infinitesimal jackknife.
\newblock \emph{Journal of Machine Learning Research}, 15:\penalty0 1625--1651,
  2014.

\bibitem[Wahab et~al.(2020)Wahab, Jain, Tyrrell, Seas, Kotthoff, and
  Johnson]{Wahab:20}
Hud Wahab, Vivek Jain, Alexander~Scott Tyrrell, Michael~Alan Seas, Lars
  Kotthoff, and Patrick~Alfred Johnson.
\newblock Machine-learning-assisted fabrication: Bayesian optimization of
  laser-induced graphene patterning using in-situ raman analysis.
\newblock \emph{Carbon}, 167:\penalty0 609--619, 2020.

\bibitem[Wang et~al.(2018)Wang, Gehring, Kohli, and Jegelka]{Wang:18}
Z.~Wang, C.~Gehring, P.~Kohli, and S.~Jegelka.
\newblock Batched large-scale bayesian optimization in high-dimensional spaces.
\newblock \emph{Proceeding of Machine Learning Research}, 84:\penalty0 745 --
  754, 2018.

\bibitem[Ward et~al.(2018)Ward, Dunn, Faghaninia, Zimmermann, Bajaj, Wang,
  Chen, Bystrom, Dylla, Chard, Asta, Persson, Snyder, Foster, and
  Jain]{Ward:18}
L.~Ward, A.~Dunn, A.~Faghaninia, N.~E.~R. Zimmermann, S.~Bajaj, J.~H. Wang,
  Q.and~Montoya, J.~Chen, K.~Bystrom, M.~Dylla, K.~Chard, M.~Asta, K.~Persson,
  G.~J. Snyder, I.~Foster, and A.~Jain.
\newblock Matminer: An open source toolkit for materials data mining.
\newblock \emph{Computational Materials Science}, 152:\penalty0 60--69, 2018.

\bibitem[Ward et~al.(2016)Ward, Agrawal, Choudhary, and Wolverton]{Ward:16}
Logan Ward, Ankit Agrawal, Alok Choudhary, and Christopher Wolverton.
\newblock A general-purpose machine learning framework for predicting
  properties of inorganic materials.
\newblock \emph{npj Computational Materials}, 2, 2016.

\bibitem[Zhang et~al.(2020)Zhang, Zimmerman, Nettleton, and Nordman]{Zhang:20}
Haozhe Zhang, Joshua Zimmerman, Dan Nettleton, and Daniel~J. Nordman.
\newblock Random forest prediction intervals.
\newblock \emph{The American Statistician}, 74:\penalty0 392--406, 2020.

\end{thebibliography}

\newpage

\appendix
\section{Compendium of Test Problems}\label{apx:test_problems}

The sections below describe how to generate ground-truth data for each test problem. For synthetic problems additional noise may be added in the form of $\epsilon * \mathcal{N}(0, 1)$, where $\epsilon$ is the noise level.

\subsection{Tophat}\label{apx:tophat}

The tophat method used in Figure \ref{fig:jackknife_vs_error}a is defined in Equation \ref{eqn:tophat}

\begin{equation}\label{eqn:tophat}
y = 
\begin{cases}
1.0, & \text{if } |x| < 0.33\\
0.5, & \text{if } 0.33 \leq |x| < 0.67\\
0, & \text{otherwise}
\end{cases}
\end{equation}

\subsection{Cubic}\label{apx:cubic}

The cubic method used in Figure \ref{fig:jackknife_vs_error}b is simply $y = x^3$.

\subsection{Friedman-Grosse}\label{apx:Friedman_Grosse}

The Friedman-Grosse function \citep{Friedman:83}, in Equation \ref{eqn:Friedman-Grosse} is defined on the unit hypercube in at least 5 dimensions. In this paper we use 8 dimensions, but dimensions 6-8 do not contribute to the output.

\begin{equation}\label{eqn:Friedman-Grosse}
    y_0 = 10 \sin(\pi x_0 x_1) + 20 (x_2 - 0.5)^2 + 10 x_3 + 5 x_4
\end{equation}

\subsubsection{Additional Outputs for Calibration Studies}

For the purposes of investigating the prediction distribution's calibration (Figure \ref{fig:covariance_metrics}) we generate two additional outputs, both of which are correlated with $Y_0$.
The first additional output, $Y_1$, has some fixed linear correlation, $\rho$, with $Y_0$.
The second additional output, $Y_2$, varies between being positively and negatively correlated with $Y_0$.

We now describe the procedure to generate a new variable, $Y_1$, that has some fixed linear correlation with the existing output variable, $Y_0$.
Assume that we have first drawn $N$ points $\vec{x}_i$ with associated output values $y_{0i}$.
To created a new variable with known correlation we want to mix together those $y_{0i}$ with an uncorrelated signal, $Z'$.
We create an uncorrelated signal by drawing $N$ values a unit normal distribution, $z_i$, performing linear least squares regression $Z = m Y$, and computing an orthogonal signal $Z' = Z - m Y$.
Let $\sigma_{Z'}$ be the standard deviation of the residuals and let $\sigma_{Y0}$ be the standard deviation of the $Y_0$ values.
For a desired correlation coefficient $\rho$, the first additional output is $Y_1 = \rho \sigma_{Z'} Y_0 + \sqrt{1 - \rho^2} \sigma_{Y0} Z'$.

Because $Y_0$ and $Z'$ are orthogonal, $\text{Cov}(Y_1, Y_0) = \rho \sigma_{Z'} \text{Cov}(Y_0, Y_0) = \rho \sigma_{Z'} \sigma_{Y_0}^2$. Also, $\text{Var}(Y_1) = \rho^2 \sigma_{Z'}^2 \sigma_{Y_0}^2 + (1-\rho^2) \sigma_{Y_0}^2 \sigma_{Z'}^2 =  \sigma_{Y_0}^2 \sigma_{Z'}^2$. The correlation coefficient between $Y_0$ and $Y_1$ is therefore exactly equal to $\rho$.
In this work we set $\rho = 0.9$ unless otherwise specified.

The second additional output is defined with a quadratic equation. Let $\mu_{y0}$ be the mean of the values $y_{0i}$. We then define $y_{2i} = (y_{0i} - \mu_{y0})^2 + f * \mathcal{N}(0, 1)$, where $f$ is some adjustable parameter. In this work we set $f = 0.5$ unless otherwise specified, which leads to a strong (but non-uniform) correlation.

\subsubsection{Additional Output for Sequential Learning}

For the purposes of synthetic sequential learning we generate the second output differently, because the intent is to create a problem in which the correlation coefficient varies depending on the region of input space. We first generate 128 inputs by sampling uniformly from the hypercube and calculate the first output $y$ using equation \ref{eqn:Friedman-Grosse}. We then create another input, ``phase,'' and randomly assign each point to either ``A'' or ``B''. For the points in phase A, the second output is generated using the linear procedure in the previous subsection, using $\rho = 0.98$. For the points in phase B, the second output is equal to $\sqrt{30^2 - y^2}$ (30 is the largest possible value of the Friedman-Grosse function).

\begin{figure}[ht]
    \centering
    \includegraphics[scale=0.65]{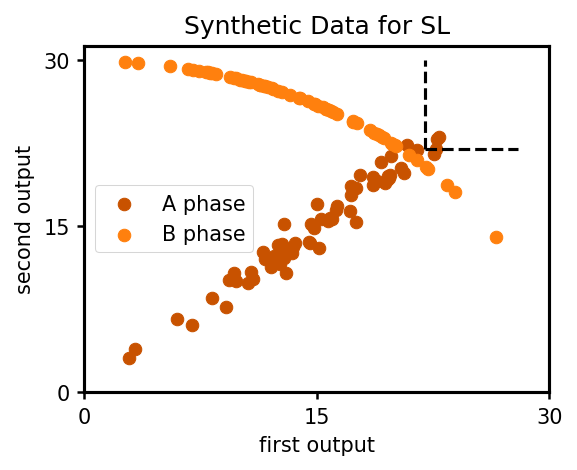}
    \caption{Synthetic data set generated using the Friedman-Grosse function for simulated sequential learning. In one region (``A phase'') there is a strong positive correlation between the outputs, while in the other region (``B phase'') there is a strong negative correlation between the outputs. The dashed lines correspond to the two objectives used in the simulated sequential learning study.}
    \label{fig:synthetic_data}
\end{figure}

This data set is plotted in Figure \ref{fig:synthetic_data} along with dashed lines at 22 for both the first and second output, corresponding to the objectives used in the SL simulation. We see that only two points satisfy both objectives.

\subsection{Friedman-Silverman}\label{apx:Friedman_Silverman}

The Friedman-Silverman function \citep{Friedman:89}, in Equation \ref{eqn:Friedman-Silverman} is defined on the unit hypercube in at least 5 dimensions. In this paper we use 12 dimensions, but dimensions 6-12 do not contribute to the output.

\begin{equation}\label{eqn:Friedman-Silverman}
    y_0 = 0.1 e^{4x_0} + \frac{4}{1 + e^{-20 (x_1 - 0.5}} + 3x_2 + 2x_3 + x_4
\end{equation}

For the purposes of investigating the prediction distribution's calibration (Figure \ref{fig:covariance_metrics}), two additional outputs are generated in the same way as for the Friedman-Grosse function.

\subsection{Thermoelectrics}\label{apx:thermoelectrics}

\begin{figure}[h]
    \centering
    \includegraphics[scale=0.65]{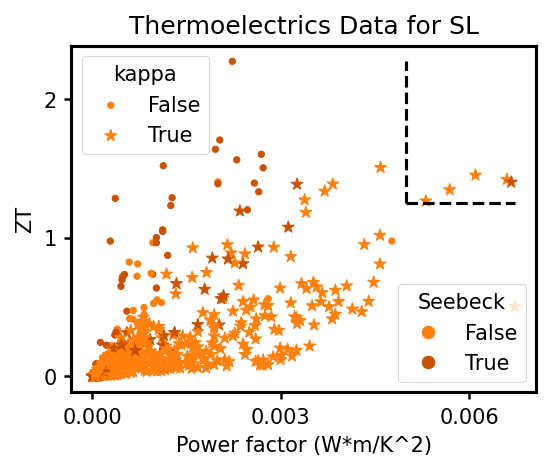}
    \caption{A projection of the thermoelectrics data set, meant to visualize the difficulty of the simluated SL problem. Dashed lines indicate the objectives for ZT (1.25) and power Factor (0.005 $\text{Wm}/\text{K}^2$). Points are colored red or orange according to whether or not the Seebeck coefficient exceeds the objective of 175 $\mu\text{V}/\text{K}$. The marker is a star or a circle depending on whether or not the thermal conductivity, kappa, exceeds the objective of 1.5 W/mK.}
    \label{fig:thermoelectrics_data}
\end{figure}

The data come from \citet{Gaultois:13}. There are 688 thermoelectric materials, each with a chemical formula, crystallinity (either polycrystalline or single crystal), and temperature at which the measurements were taken. The chemical formulae were featurzed using the Matminer package \citep{Ward:18} to calculate the Magpie features \citep{Ward:16}. Along with temperature and crystallinity, these are the inputs to the ML models. There are five measured output properties: ZT, Seebeck coefficient, thermal conductivity, power factor, and log resistivity. One projection of this data set is plotted in Figure \ref{fig:thermoelectrics_data}.

\subsection{Mechanical Properties}\label{apx:mechanical_properties}

The data come from \citet{Borg:20}.
There are 630 multi-principal element alloys, some measured under a variety of conditions.
The data set contains several inputs and several outputs, many of which are sparse.
To yield a dense training table we only consider the following inputs: processing method (cast, wrought, anneal, powder, and other), crystal structure (BCC, FCC, and other), test type (compression or tension), and chemical formula.
The chemical formulae are featurized as with the thermoelectrics data. We only consider two output properties: Young's Modulus and Elongation.

We only keep rows that have values for all inputs and outputs. We only keep rows that were measured at room temperature (between 20 and 25${}^o$C). If there are multiple rows with identical inputs, we average their output properties. This yields 287 rows.

\section{Analysis of the Recalibrated Bootstrap}\label{apx:recalibrated_bootstrap}

A decision tree partitions the domain into rectangular subspaces, each corresponding to a terminal node of the tree. The prediction in this subspace is the average of the values of the training data that appear in that node. This training data $\rightarrow$ subspace correspondence is a complicated function of the training data, the splitting function, and random variables (such as which input dimensions are being considered), but we can write the tree's predictions as a weighted sum of the training data. The RF prediction is therefore given by equation \ref{eqn:weighted_RF}, where $\{Z_n\} \in Z$ is represents the training data draw of size $n$ and $\phi \in \Phi$ represents a random variables that controls the splitting. The weights $w_{bi}(\vec{x})$ are non-zero only when $\vec{x}_i$ is ``close'' to $\vec{x}$, and the weights sum to 1 for each tree and value of $\vec{x}$.

\begin{equation}\label{eqn:weighted_RF}
    \hat{\theta}(\vec{x}) = \frac{1}{B} \sum_{b=1}^B \sum_{i=1}^N w_{bi} (\vec{x}; \{Z_n\}, \phi) y_i
\end{equation}

The out-of-bag prediction for $\vec{x}_i$ is similar, but the sum is taken only over the trees for which $\vec{x}_i$ is not present in the bootstrap sample. This is, on average, $B/e$ trees.

\begin{equation}\label{eqn:weighted_RF_oob}
    \hat{\theta}_{(-i)}(\vec{x}_i) = \frac{1}{B/e} \sum_{b:z_i \notin Z_b} \sum_{j} w_{bj} (\vec{x}_i; \{Z_n\}, \phi) y_i
\end{equation}

\subsection{Asymptotic Equivalence of Standard Residual Distributions}\label{apx:residual_equivalence}

We wish to show that the distribution of true standard residuals is equivalent to the distribution of OOB standard residuals:

\begin{equation}
    \Big\langle \frac{\hat{\theta}_{(-i)}(\vec{x}_i) - y_i}{\hat{s}_{(-i)}(\vec{x}_i)} \Big\rangle_{\{Z_n\}, \phi, z_j} \overset{?}{=} \Big\langle \frac{\hat{\theta}(\vec{x}) - (f(\vec{x}) + \epsilon \mathcal{N}(0, 1)}{\hat{s}(\vec{x})} \Big\rangle_{\{Z_n\}, \phi, x, y}
\end{equation}

The expectation is taken over all training data draws $\{Z_n\}$ and parameters $\phi$, and on the LHS over all training data ($\vec{x}_j, y_j) \in \{Z_n \backslash z_i\}$ while on the RHS it is over all $\vec{x}$ in the domain and all associated observations drawn from $y = f(\vec{s}) + \epsilon \mathcal{N}(0, 1)$.

Consider what a fixed point $\vec{x}$ contributes to each side of this equation. Because the labels $y_i$ are generated according to $f(\vec{x}) + \epsilon \mathcal{N}(0, 1)$, and drawing the noise at $\vec{x}$ is independent of $\hat{s}(\vec{x})$, we can replace both $y_i$ and $f(\vec{x}) + \epsilon \mathcal{N}(0, 1)$ in expectation with $f(\vec{x})$. On the LHS we therefore have the mean standard residual at $\vec{x}$ for an RF model trained with $N-1$ iid training points and $B/e$ trees. On the right we have the mean standard residual at $\vec{x}$ for an RF model trained with $N$ iid training points and $B$ trees. For large values of $B$ and $N$, these are equivalent.

\subsection{Bootstrap Variance is Sensitive to the True Error}\label{apx:recalibrated_bootstrap_captures_error}

The total expected squared error due to a model can be represented as the sum of the bias squared, the model variance, and the noise variance. We wish to show that the bootstrap variance (and hence the recalibrated bootstrap standard deviation) is sensitive to each of these terms.

\subsubsection{Noise Variance}

The bootstrap variance is written in Equation \ref{eqn:bootstrap_variance}

\begin{equation}\label{eqn:bootstrap_variance}
    \langle \hat{s}^2(\vec{x}) \rangle = \Big\langle \frac{1}{B-1} \sum_b (t_b(\vec{x}) - \hat{\theta}(\vec{x}))^2 \Big\rangle_{\{Z_n\}, \phi}
    = \frac{B}{B-1} \langle t_b(\vec{x})^2 \rangle_{\{Z_n\}, \phi, b} - \frac{B}{B-1} \langle \hat{\theta}(\vec{x})^2 \rangle_{\{Z_n\}, \phi}
\end{equation}

The simplification is possible because $\hat{\theta}$ is the mean of $t_b$. First, we consider the expectation of $t_b(\vec{x})^2$:

\begin{equation}
    \langle t_b(\vec{x})^2 \rangle_{\{Z_n\}, \phi, b} = \Big\langle \Big( \sum_i w_{bi} (\vec{x}; \{ Z_n \}, \phi) (f(\vec{x}_i) + \epsilon \mathcal{N}(0, 1)) \Big)^2 \Big\rangle_{\{Z_n\}, \phi, b}
\end{equation}

The weights $w_{bi}(\vec{x})$ mostly depend on which training data are ``close'' to $\vec{x}$ in input space. There is some dependence on the exact values of $y_i$, but we can consider the weights $w_{bi}$ to be roughly independent of the noise drawn from $\epsilon \mathcal{N}(0, 1)$. In that case the expectation simplifies to Equation \ref{eqn:tree_variance_noise}, where $L$ is the typical number of unique training observations on a leaf node.

\begin{equation}\label{eqn:tree_variance_noise}
    \langle t_b(\vec{x})^2 \rangle \approx \Big\langle \Big( \sum_i w_{bi} (\vec{x}; \{ Z_n \}, \phi) f(\vec{x}_i)\Big)^2 \Big\rangle_{\{Z_n\}, \phi, b} + \frac{\epsilon^2}{L}
\end{equation}

Similarly, $\langle \hat{\theta}(\vec{x})^2 \rangle_{\{Z_n\}, \phi}$ also has a term that goes as $\epsilon^2$. But since there is an additional average over the bootstrap samples and each bootstrap sample masks off some training data, there will be a larger number of independent noise draws being averaged together and hence the effective value of $L$ is larger. Therefore, $\langle \hat{s}(\vec{x})^2 \rangle$ has a term that is proportional to the noise variance $\epsilon^2$. This term is smaller if there are more training examples on each leaf node, but that makes sense since we would expect a model that averages over more training data to be less sensitive to the noise.

In Appendix D we show that, as the noise becomes the dominant term, the recalibrated bootstrap correctly identifies each observation as being an independent random variable with uncertainty equal to the noise.

\subsubsection{Model Variance}

Having considered the noise, we set $\epsilon = 0$ for the remainder of this Appendix. We next consider the variance due to the finite training data set size and the parameters of the model, $\langle \hat{\theta}(\vec{x})^2 \rangle_{\{Z_n\}, \phi} - \langle \hat{\theta}(\vec{x}) \rangle^2_{\{Z_n\}, \phi}$. Assume for now that there is no bias, so $\langle \hat{\theta}(\vec{x}) \rangle = f(\vec{x})$, and without loss of generality assume $f(\vec{x}) = 0$. Our task is to show that $\langle \hat{s}(\vec{x})^2\rangle \propto \langle \hat{\theta}(\vec{x})^2\rangle_{\{Z_n\}, \phi}$, which by Equation \ref{eqn:bootstrap_variance} is equivalent to showing that $\langle t_b(\vec{x})^2 \rangle_{\{Z_n\}, \phi, b} \propto \langle \hat{\theta}(\vec{x})^2\rangle_{\{Z_n\}, \phi}$.

But this is clearly true because $\hat{\theta}(\vec{x})$ is the mean of $t_b(\vec{x})$. Consider some fixed $\{Z_n\}$ and $\phi$. The $t_b(\vec{x})$ are the predictions made by all trees trained on bootstrap replicates of $\{Z_n\}$. The $\hat{\theta}(\vec{x})$ are the averages of all sets of $B$ draws from the distribution of $t_b(\vec{x})$. By the law of large numbers, the variance of the latter is equal to the variance of the former times $1/B$. Hence, $\langle \hat{s}(\vec{x})^2\rangle$ is sensitive to model variance.

\subsubsection{Bias Squared}

Assume for simplicity that the training data are independent and identically distributed, that $f(\vec{x})$ is continuous, and that the random forest is full-depth. In this case the model is asymptotically consistent and free of bias \textit{except} for at the boundaries. As a way of illustrating how the bootstrap variance can pick up bias, we consider the boundary of a one-dimensional test problem. A more rigorous investigation would need to consider multi-variate and non-identically distributed data, which is the case for the real-world data sets used here.

Let the domain be $[0, 1]$ and let the function $f(x)$ have slope $m$ at $x=0$. Because the trees are grown to full depth the predicted value at $x=0$ will be equal to the value of the training point with the smallest value of $x$. Because the training data are drawn uniformly this is equal in expectation to $1/(N+1)$. Assuming the linear approximation of $f(x)$ is valid over the length-scale of $1/(N+1)$, the predicted value will therefore be equal in expectation to $m/(N+1)$.

However, roughly $1/e$ trees will be missing this data point. Most of these trees will make a prediction that is equal to the value of the training point with the second-smallest value of $x$, which in expectation is at $2/(N+1)$ and has a value of $2m/(N+1)$. There are other terms as well, but they are all proportional to $m/(N+1)$. Hence the bias goes as $m/(N+1)$ and the bias squared goes as $m^2/(N+1)^2$.

Now consider the bootstrap variance, which is the mean of $(t_b(0) - \hat{\theta}(0))^2$. As discussed above, $t_b(0)$ is always proportional to $m/(N+1)$. Hence $\hat{\theta}(0) \propto m/(N+1)$. Though the sum is complex, every term has a factor of $m/(N+1)$ that can be factored out, and so the bootstrap variance goes as $m^2/(N+1)^2$, just like the bias squared does.

\subsubsection{Conclusion}

These arguments are not precise, and in fact we do not expect the bootstrap variance to perfectly capture the true residual. We merely show that it has the capability to pick up on the bias, the variance, and the noise. Its efficacy and ability to balance these three terms is shown through the numerical experiments in this manuscript.

\section{Details of Jackknife Variance and Covariance (Including the Bias-Correction Term)}\label{apx:jackknife}

\citet{Wager:14} introduces bias-corrected versions of two methods to estimate a confidence interval: the Infinitesimal Jackknife (IJ) and the Jackknife after Bootstrap (JaB). Here we generalize those derivations from variance to covariance. \citet{Wager:14} found that the IJ and JaB had opposite lowest-order biases, and hence suggested averaging them. In this manuscript, whenever ``jackknife methods'' are invoked, it is the average of the IJ and the JaB (co)variance. The square root of this term is referred to as the ``jackknife standard deviation.''

Although the derivations below consider covariance between two outputs predicted by one ensemble model at one input point, they also apply to the covariance between predictions made at two distinct input points or to predictions made by two distinct ensemble models. Indeed, \citet{Ghosal:21} derive an analogue of Equation \ref{eqn:IJ_cov_full} but their focus is on the covariance between the predictions made by two models, in order to test if the models are equivalent.

\subsection{Infinitesimal Jackknife Covariance}

We show that the bias-corrected IJ covariance between outputs $j$ and $k$ at point $\vec{x}$ is given by Equation \ref{eqn:IJ_cov_full}, where $Y_{bi}$ is the number of times training datum $i$ appears in bag $b$.

\begin{equation}\label{eqn:IJ_cov_full}
\begin{aligned}
    \text{Cov}_{IJ}[j, k](\vec{x}) \approx & \sum_{i=1}^N \Big( \sum_{b=1}^B \frac{(Y_{bi} - 1) (t_{bj}(\vec{x}) - \hat{\theta}_j(\vec{x}))}{B} \Big) \Big( \sum_{b=1}^B \frac{(Y_{bi} - 1) (t_{bk}(\vec{x}) - \hat{\theta}_k(\vec{x}))}{B} \Big)\\
    & - \frac{N-1}{B} \sum_{b=1}^B (t_{bj}(\vec{x}) - \hat{\theta}_j(\vec{x})) (t_{bk}(\vec{x}) - \hat{\theta}_k(\vec{x}))
\end{aligned}
\end{equation}

\subsubsection{Main Term}

We follow the arguments in \citet{Efron:14}, which themselves are similar to those in \citet{Efron:82}, but we generalize to covariance. Consider two estimators, $\hat{\theta}_j$ and $\hat{\theta}_k$, that evaluate some functions $\theta_j$ and $\theta_k$ for a given distribution of training data. For our purposes, $\theta_j$ and $\theta_k$ correspond to two outputs that the model predicts. The predictions depend on the bootstrap samples, which are generated by drawing $N$ samples from $N$ training data according to some probability vector, $\vec{p}$ of length $N$. For an ordinary bootstrap, $\vec{p} = \vec{p}_0$, a uniform vector for which each entry is equal to $1/N$. Equivalently, this can be thought of as an isotropic rescaled multinomial distribution, $\mathbb{M}$.

We are therefore interested in the covariance between $\theta_j(\vec{p})$ and $\theta_j(\vec{p})$ over the multinomial distribution. Both $\theta_j$ and $\theta_k$ are effectively functions of $\vec{p}$, and we linearize them around $\vec{p}_0$ as $\theta_j(\vec{p}) = \theta_j(\vec{p}_0) + (\vec{p} - \vec{p}_0) \cdot \vec{U}$, where $\vec{U}$ is the influence function. Similarly, let $\vec{V}$ be the influence function of $\theta_j(\vec{p})|_{\vec{p}_0}$. Because $\sum_i U_i = \sum_i V_i = 0$, the covariance reduces as shown below.

\begin{equation*}
\begin{aligned}
    \text{Cov}_{\mathbb{M}}[\theta(\vec{p}), \phi(\vec{p})] & = \mathbb{E}_{\mathbb{M}}[(\theta(\vec{p}) - \bar{\theta}(\vec{p}))(\phi(\vec{p}) - \bar{\phi}(\vec{p}))]\\
    & = \mathbb{E}_{\mathbb{M}}[(\vec{p} - \vec{p}_0) \cdot \vec{U} (\vec{p} - \vec{p}_0) \cdot \vec{V})] \\
    & = \mathbb{E}_{\mathbb{M}}[(\vec{p} \cdot \vec{U}) (\vec{p} \cdot \vec{V})] \\
    & =\mathbb{E}_{\mathbb{M}}\Big[ \sum_{i=1}^N p_i^2 U_i V_i + \sum_{i=1}^N \sum_{l \neq i} p_i p_l U_i V_l \\
    & = \sum_{i=1}^N U_i V_i \mathbb{E}_{\mathbb{M}}[p_i^2] + \sum_{i=1}^N \sum_{l \neq i} U_i V_l \mathbb{E}_{\mathbb{M}} [ p_i p_l]
\end{aligned}
\end{equation*}

The expectation values of $p_i^2$ and $p_i p_j$ are those of $\mathbb{M}$, for which the mean of each term is $1/N$, the variance of each term is $1/N^2 - 1/N^3$, and the covariance between any pair of terms is $-1/N^3$. This yields the following.

\begin{equation*}
    \text{Cov}_{\mathbb{M}}[\theta_j(\vec{p}), \theta_k(\vec{p})] = \frac{1}{N^2} \sum_{i=1}^N U_i V_i
\end{equation*}

What are $U_i$ and $V_i$? By taking the derivative, \citet{Efron:14} derives them to be equal to $N \text{Cov}_*[Y_{bi}, t_b(\vec{x})]$. The asterisk indicates that the covariance is taken over the bootstrap samples. The IJ covariance estimate between outputs $j$ and $k$ is therefore given by Equation \ref{eqn:IJ_cov_main_term}, which is a natural generalization of the IJ variance.

\begin{equation}\label{eqn:IJ_cov_main_term}
    \text{Cov}_{IJ}[j, k](\vec{x}) = \sum_{i=1}^N \text{Cov}_*[Y_{bi}, t_{bj}(\vec{x})] \text{Cov}_*[Y_{bi}, t_{bk}(\vec{x})]
\end{equation}

When expanding this covariance explicitly we need to know the mean values of $Y_{bi}$ and $t_b$, which are 1 and $\hat{\theta}$.

\subsubsection{Bias Correction Term}

Equation \ref{eqn:IJ_cov_main_term} is exact as $B \to \infty$, but in practice we only consider a subset of bags. Let $A_j^B = \text{Cov}_*[Y_{bi}, t_{bj}(\vec{x})]$ and $A_j^\infty$ be the same for infinite $B$. One term of the IJ sum is, in expectation:

\begin{equation*}
\begin{aligned}
    \mathbb{E}_*[A_j^B A_k^B] & = \mathbb{E}_*[A_j^B] \mathbb{E}_*[A_k^B] + \text{Cov}_*[A_j^B, A_k^B]\\
    & = A_j^\infty A_k^\infty + \text{Cov}[A_j^B, A_k^B]
\end{aligned}
\end{equation*}

The difference between the finite-B value and the ideal value is therefore given by $-N \text{Cov}_*[A_j^B, A_k^B]$. Writing this out in detail we get Expression \ref{eqn:IJ_bias_correction_partial}, where the expectation is now taken over the training data.

\begin{equation}\label{eqn:IJ_bias_correction_partial}
    -\frac{N}{B} \mathbb{E} \Big[ \text{Cov}_*[Y_{bi} * (t_{bj}(\vec{x})) - \hat{\theta}_j(\vec{x}), Y_{bi} * (t_{bk}(\vec{x}) - \hat{\theta}_j(\vec{x}))] \Big]
\end{equation}

As in \citet{Wager:14} we treat $Y_{bi}$ and $t_b$ as independent, meaning we can pull out the $Y_{bi}$ terms into their own variance term $\text{V}_*[Y_{bi}]$, which is equal to the variance of a multinomial distribution, or $(N - 1) / N$. The remaining covariance between the trees $t_{bj}$ and $t_{bk}$ is written out explicitly, and we arrive at the following bias-correction term.

\begin{equation}\label{eqn:IJ_bias_correction}
    - \frac{N-1}{B} \sum_{b=1}^B (t_{bj}(\vec{x}) - \hat{\theta}_j(\vec{x})) (t_{bk}(\vec{x}) - \hat{\theta}_k(\vec{x}))
\end{equation}

Note that this differs slightly from the bias-correction in \citet{Wager:14}, because they mistakenly set the variance of $Y_{bi}$ equal to 1 and hence have a factor of $N$ instead of $N-1$ (this occurs right before equation 11 in \citet{Wager:14}).

Combining Equations \ref{eqn:IJ_cov_main_term} and \ref{eqn:IJ_bias_correction} we arrive at Equation \ref{eqn:IJ_cov_full}

\subsection{Jackknife after Bootstrap Covariance}

We show that the bias-corrected JaB covariance between outputs $j$ and $k$ at point $\vec{x}$ is given by Equation \ref{eqn:JaB_cov_full}.

\begin{equation}\label{eqn:JaB_cov_full}
\begin{aligned}
    \text{Cov}_{JaB}[j, k](\vec{x}) \approx & \frac{N-1}{N} \sum_{i=1}^N \Big( \hat{\theta}_{j(-i)}(\vec{x}) - \hat{\theta}_j(\vec{x}) \Big) \Big( \hat{\theta}_{k(-i)}(\vec{x}) - \hat{\theta}_k(\vec{x}) \Big)\\
    & - (e - 1) \frac{N-1}{B} \sum_{b=1}^B (t_{bj}(\vec{x}) - \hat{\theta}_j(\vec{x})) (t_{bk}(\vec{x}) - \hat{\theta}_k(\vec{x}))
\end{aligned}
\end{equation}

\subsubsection{Main Term}

The derivation of the JaB in \citet{Efron:82} is identical to that of the IJ, but the influence function is different. The generalization to covariance therefore works out the same way and we get Equation \ref{eqn:JaB_cov_derivation}.

\begin{equation}\label{eqn:JaB_cov_derivation}
    \text{Cov}_{JaB}[j, k](\vec{x}) = \frac{N-1}{N} \sum_{i=1}^N \Big( \hat{\theta}_{j(-i)}(\vec{x}) - \hat{\theta}_j(\vec{x}) \Big) \Big( \hat{\theta}_{k(-i)}(\vec{x}) - \hat{\theta}_k(\vec{x}) \Big)
\end{equation}

\subsubsection{Bias Correction Term}

The derivation of the JaB bias term is the same as it is for the IJ, above, but instead of $A_{j}^B$ the relevant term for some training datum $i$ is now $\hat{\Delta}_{ij} \equiv \hat{\theta}_j^B - \hat{\theta}_{(-i)j}^B$ (the dependence on $\vec{x}$ is elided for simplicity) and there is also an overall factor of $(N-1)/N$. In order to evaluate $\text{Cov}_*[\hat{\Delta_{ij}}, \hat{\Delta_{ij}}]$ we follow Appendix A of \citet{Wager:14} but generalize from variance to covariance.

Using the law of total covariance, we can expand this term out as follows:

\begin{equation*}
    \text{Cov}_*[\hat{\Delta}_{ij}, \hat{\Delta}_{ik}] = \mathbb{E}_*\Big[ \text{Cov}_*[\hat{\Delta}_{ij} | B_i, \hat{\Delta}_{ik} | B_i] \Big] + \text{Cov}_* \Big[ \mathbb{E}_*[\hat{\Delta}_{ij} | B_i], \mathbb{E}_*[\hat{\Delta}_{ik} | B_i]\Big]
\end{equation*}

The ``covariance of expectation'' term simplifies just as the analogous ``variance of expectation'' term in \citet{Wager:14}, to $\Delta_{ij} \Delta_{ik} \mathcal{O}(1/B)$, where the $\Delta$ terms without a hat indicate the average over all possible bootstrap samples.

To evaluate the ``expectation of covariance'' term we re-write $\hat{\Delta}$ as a sum over $B_i$ bags that do not contain point $i$ and a subsequent sum over $B-B_i$ bags that do contain point $i$. On average, $B_i = B/e$.

\begin{equation*}
    \hat{\Delta}_{ij} = \frac{1}{B_i} \sum_{b=1}^{B_i} t_{bj} - \frac{1}{B} \sum_{b=1}^B t_{bj} = \frac{B - B_i}{B B_i} \sum_{b=1}^{B_i} t_{bj} - \frac{1}{B} \sum_{b=B_i+1}^B t_{bj}
\end{equation*}

The covariance between $\hat{\Delta}_{ij}$ and $\hat{\Delta}_{ik}$ then breaks out into the sum of many covariance terms, all of the form $\text{Cov}[t_{bj}, t_{b'k}]$. But because the bags are independent, all of these terms are 0 unless $b=b'$. We therefore have $B_i$ instances of $\text{Cov}[t_{bj} | B_i = 0, t_{bk} | B_i = 0]$ and $B-B_i$ instances of $\text{Cov}[t_{bj} | B_i \neq 0, t_{bk} | B_i \neq 0]$, which we denote $cov_i^{(0)}$ and $cov_i^{(+)}$.

\begin{equation}\label{eqn:JaB_bias_derivation}
    \text{Cov}_*[\hat{\Delta}_{ij} | B_i, \hat{\Delta}_{ik} | B_i] = \Big( \frac{B-B_i}{BB_i} \Big)^2 B_i cov_i^{(0)} + \frac{1}{B^2} (B-B_i) cov_i^{(+)}
\end{equation}

Equation \ref{eqn:JaB_bias_derivation} is identical to the analogous equation in \citet{Wager:14} except that we have the covariance between $t_{bj}$ and $t_{bk}$ instead of the variance of $t_b$. Calculating the expectation therefore proceeds the same way and we wind up with, to lowest order, $\frac{e-1}{B} \text{Cov}_*[t_{bj}, t_{bk}]$. Including the factor of $(N-1)/N$ and the sum over $N$, we wind up with expression \ref{eqn:JaB_bias_term}.

\begin{equation}\label{eqn:JaB_bias_term}
    -(e-1)\frac{N-1}{B} \sum_{b=1}^B (t_{bj}(\vec{x}) - \hat{\theta}_j(\vec{x})) (t_{bk}(\vec{x}) - \hat{\theta}_k(\vec{x}))
\end{equation}

Note that this is again slightly different from \citet{Wager:14}, who convert the factor of $N-1$ into a factor of $N$ for unclear reasons.

Combining Equations \ref{eqn:JaB_cov_derivation} and \ref{eqn:JaB_bias_term} we arrive at Equation \ref{eqn:JaB_cov_full}.

\section{More Numerical Experiments}

\subsection{The Jackknife Underestimates the True Model Error}\label{apx:jackknife_underestimates_error}

\begin{figure}[ht]
    \centering
    \includegraphics[width=0.85\linewidth]{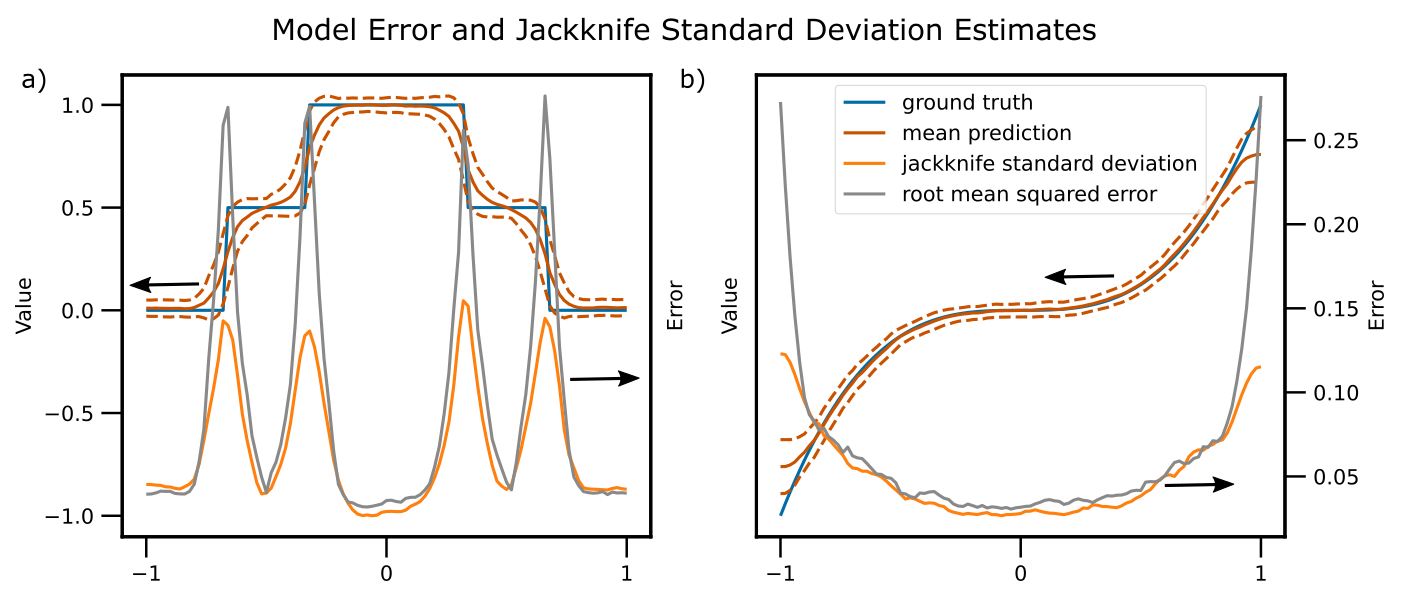}
    \caption{Jackknife standard deviation and true model root mean squared error (RMSE) for two one dimensional test functions: a) tophat and b) cubic. The ground truth and predicted value (with $\pm 1$ standard deviation) are plotted on the left axis. The jackknife standard deviation and the RMSE are both plotted on the right axis.}
    \label{fig:jackknife_vs_error}
\end{figure}

We consider two one-dimensional test problems, a double tophat and a cubic on the domain [-1, 1] (see Appendix \ref{apx:test_problems} for details). We draw 64 training data uniformly, train an RF model, and calculate its predictions on 100 points evenly spaced throughout the domain. We also calculate the jackknife standard deviation (see Appendix \ref{apx:jackknife}) and the squared error at these points. This is averaged over 250 trials. The results, in Figure \ref{fig:jackknife_vs_error}, show that a jackknife-based prediction interval can be highly over-confident, in particular when the model is biased.

\subsection{Performance on Imbalanced Data}\label{apx:imbalanced_data}

Table \ref{tbl:imbalanced_uncertainty_accuracy} shows univariate uncertainty metrics when the training and test sets are drawn from highly imbalanced distributions. We use the Young's Modulus output of the Mechanical Properties data set. In these data there are points that were measured under tension and points that were measured under compression. The training set consists of 60 tension points and 4 compression points. The test set consists of 32 compression points. We calculate prediction interval metrics for both the recalibrated bootstrap method and the method of \citet{Zhang:20}, which uses the out-of-bag residuals to compute an unconditional prediction interval. We refer to this method as "OOB constant." 50 trials were run, each involving a different draw of training/test data. Models were trained with 64 bags. Error bars are one standard error.

We see that the OOB constant method is highly over-confident -- the true residual is on average more than twice as large as the 1-$\sigma$ error bar. The recalibrated bootstrap is also over-confident but significantly less so. Interestingly the NLPD values are equivalent, highlighting the fact that NLPD must be considered as one of several metrics. The recalibration method of \citet{Palmer:21} essentially minimizes NLPD, and hence may also have trouble with imbalanced data.

\begin{table}[ht]
\caption{Comparison of the ``recalibrated bootstrap'' and ``out-of-bag constant'' prediction interval methods on a problem for which the training and test sets are drawn from different distributions.}
\label{tbl:imbalanced_uncertainty_accuracy}
\centering
\begin{tabular}{c c c c}
  & NLPD & Standard RMSE & Standard Confidence \\
 \midrule
 Out-Of-Bag Constant & 7.37 $\pm$ 0.06 & 2.12 $\pm$ 0.07 & 0.38 $\pm$ 0.02 \\
 \midrule
 Recalibrated Bootstrap & 7.39 $\pm$ 0.02 & \textbf{1.35 $\pm$ 0.07} & \textbf{0.61 $\pm$ 0.03} \\
 \bottomrule
\end{tabular}
\end{table}

\subsection{Recalibration Factors}\label{apx:recalibration_factors}

\begin{figure}[h]
    \centering
    \includegraphics[scale=0.45]{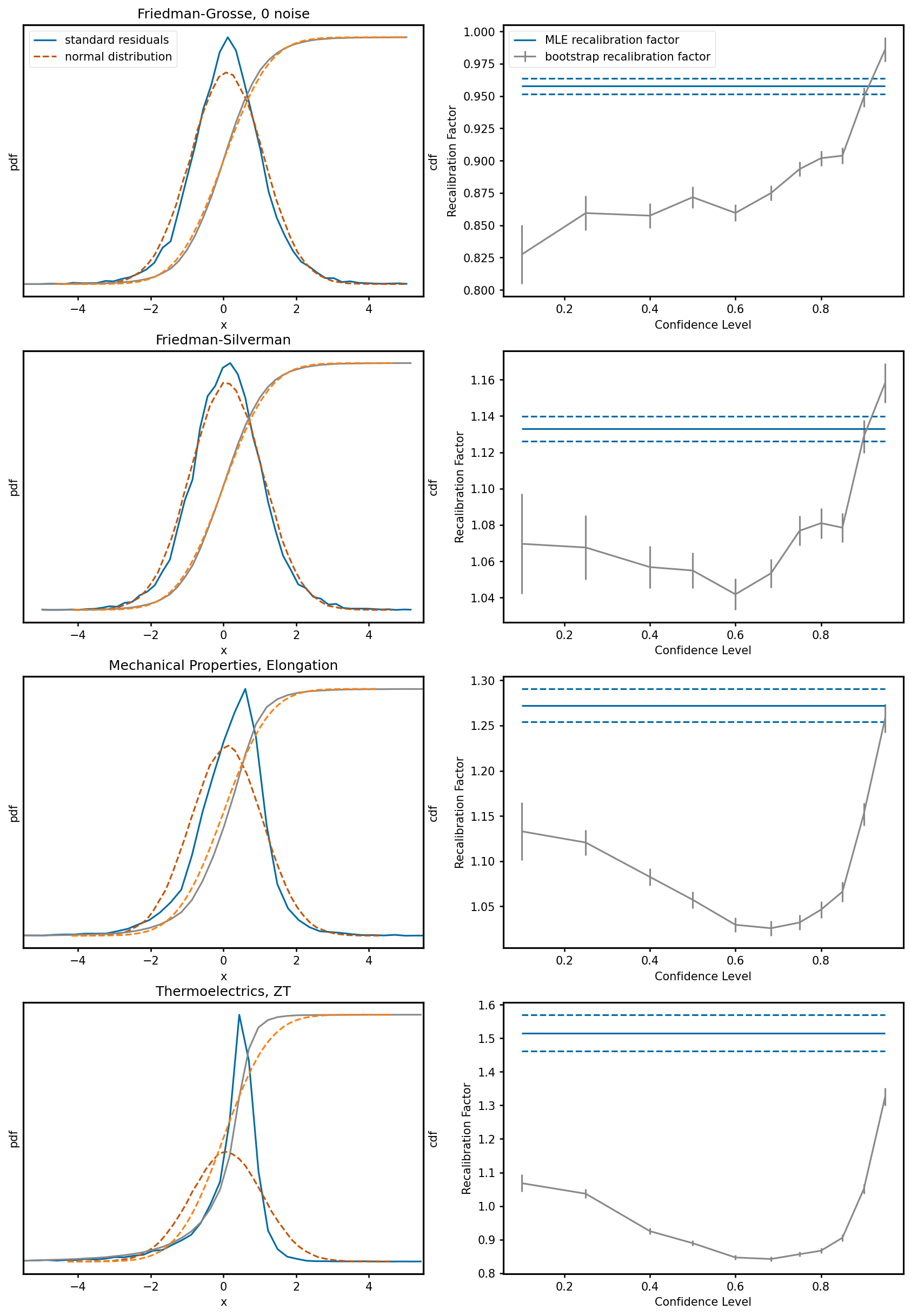}
    \caption{Investigation of the standard OOB residuals and recalibration factors for several test problems.}
    \label{fig:recalibration_supplementary}
\end{figure}

Figure \ref{fig:recalibration_supplementary} shows the distribution of standard residuals and recalibration factors (as in Figure \ref{fig:recalibration}) for four more test problems: Friedman-Grosse without noise, Friedman-Silverman, the Elongation output of the Mechanical Properties data set, and the ZT output of the Thermoelectrics data set.

We see that the other synthetic problems are well calibrated but also have some high-residual outliers. The real-world data sets are skewed and more sharply peaked.

\subsection{Prediction Interval Metrics}\label{apx:prediction_interval_metrics}

\begin{figure}[ht]
    \centering
    \includegraphics[scale=0.7]{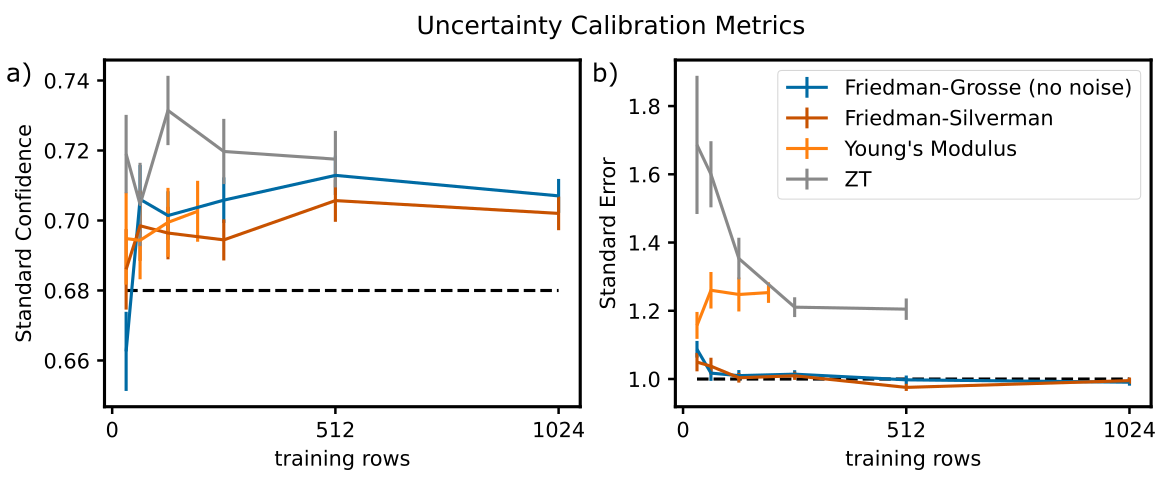}
    \caption{Univariate prediction interval metrics for several test problems. Synthetic problem have 128 test points; real data test problems have 64 test points. All results are averaged over 64 trials. Each trial involves a different set of training and test data. Error bars are 1 standard error.}
    \label{fig:univariate_metrics_supplementary}
\end{figure}

Figure \ref{fig:univariate_metrics_supplementary} shows the univariate prediction interval metrics (as in Figure \ref{fig:uncertainty_metrics} for four more test problems: Friedman-Grosse without noise, Friedman-Silverman with noise of magnitude 2.0, the Young's Modulus output of the mechanical properties data set, and the ZT output of the thermoelectrics data set.

\begin{figure}[ht]
    \centering
    \includegraphics[scale=0.6]{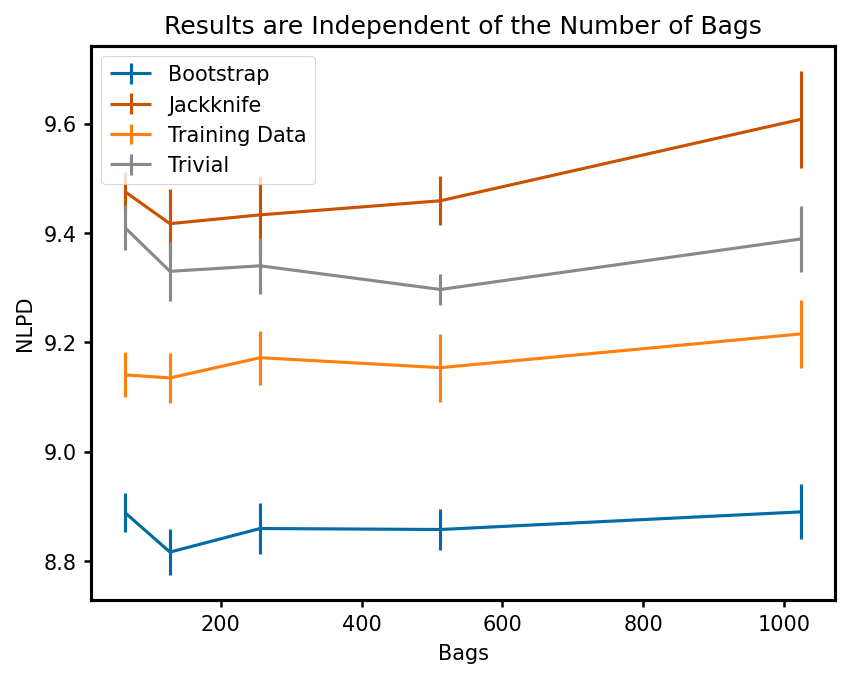}
    \caption{Increasing the number of bags does not change the relative accuracy of the correlation estimation methods. This test is on the Friedman-Grosse function with noise level 1.0, 128 training and test points, and 16 trials. Error bars are one standard error.}
    \label{fig:metrics_vary_bags}
\end{figure}

Figure \ref{fig:metrics_vary_bags} shows NLPD for the multivariate Friedman-Grosse test problem with 128 test and training points, varying the number of bags. The number of bags is largely irrelevant. The fact that a small, constant number of bags suffices to create a well-calibrated prediction interval is one of the benefits of the recalibrated bootstrap method.

\begin{figure}[ht]
    \centering
    \includegraphics[scale=0.6]{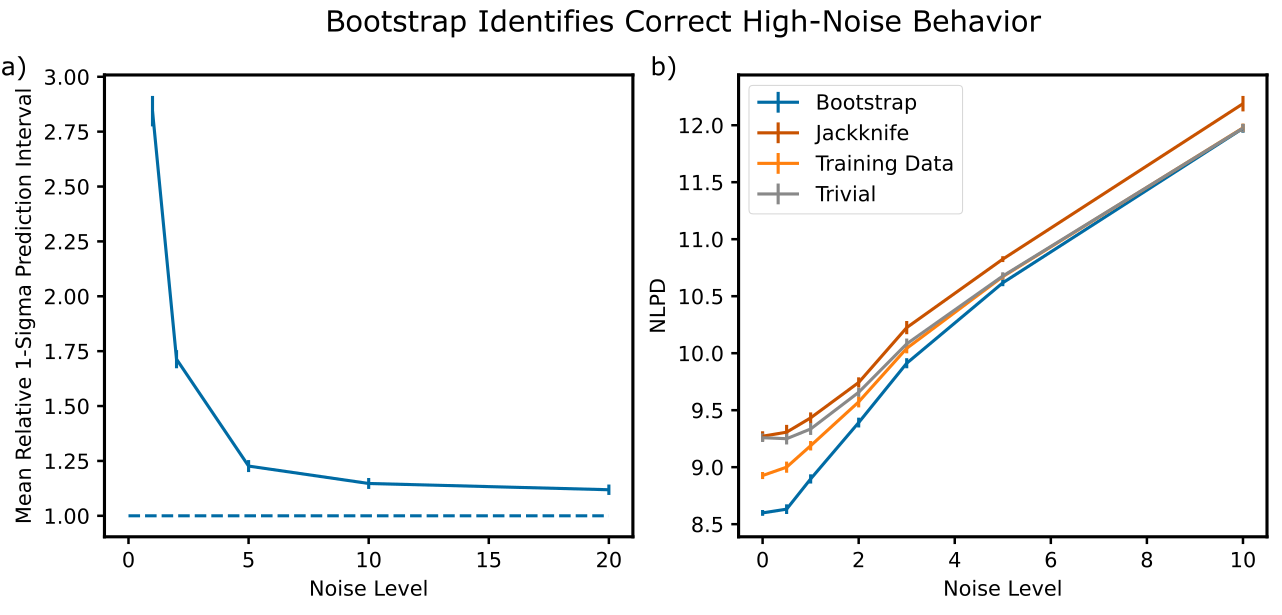}
    \caption{a) Recalibrated bootstrap prediction interval and b) NLPD as the noise level varies for the Friedman-Grosse test problem. There are 128 training points, 64 bags, and 100 trials. Error bars are one standard error.}
    \label{fig:metrics_vary_noise}
\end{figure}

Figure \ref{fig:metrics_vary_noise} shows the effect of varying the noise level for the Friedman-Grosse test problem. For a single output we consider the size of the mean 1-$\sigma$ prediction interval generated by the recalibrated bootstrap, divided by the noise level. Ideally this would approach 1 as the noise becomes dominant. Instead it approaches 1.1. The standard residual (not shown) approaches 0.9, so we see that while the recalibrated bootstrap largely picks up on the noise it is not quantitatively exactly correct.

For the multi-output problem we see that the recalibrated bootstrap and trivial methods of estimating correlation converge to the same result in the high-noise limit. Since the noise is generated independently, the correlation coefficients of the prediction distribution should be uniformly 0 in this limit. The trivial method is therefore correct, and the fact that the recalibrated bootstrap method converges to the same result indicates that it is hitting upon this correct answer as well.

\end{document}